\RequirePackage{fix-cm}
\documentclass[smallextended]{svjour3}       
\smartqed  
\usepackage[colorlinks = true,
            linkcolor = black,
            urlcolor  = blue,
            citecolor = blue,
            anchorcolor = blue]{hyperref}
\usepackage{graphicx}
\usepackage{subfig}
\usepackage{enumitem}

\usepackage{amsmath, amssymb, url, color, tikz}
\usepackage[numbers,sort]{natbib}
\usepackage{upquote}
\usepackage{mathptmx}      
\usepackage{breakcites} 
\usepackage[margin=1in]{geometry}

\newcommand{\rev}[1]{{\color{black}#1}}

\usepackage{etoolbox} 

\robustify{\texttt}

\begin{document}

\title{ABBA: Adaptive Brownian bridge-based symbolic aggregation of time series}

\author{Steven Elsworth         \and
             Stefan G\"{u}ttel}

\institute{Steven Elsworth \at
              Department of Mathematics, The University of Manchester, M13 9PL, Manchester, UK \\
              \email{steven.elsworth@manchester.ac.uk}
           \and
           Stefan G\"{u}ttel \at
           Department of Mathematics, The University of Manchester, M13 9PL, Manchester, UK \\
              \email{stefan.guettel@manchester.ac.uk}
}

\maketitle

\begin{abstract}
A new symbolic representation of time series, called ABBA, is introduced. It is based on an adaptive polygonal chain approximation of the time series into a sequence of tuples, followed by a mean-based clustering to obtain the symbolic representation. We show that the reconstruction error of this representation can be modelled as a random walk with pinned start and \rev{end points}, a so-called Brownian bridge. This insight allows us to make ABBA essentially parameter-free, except for the approximation tolerance which must be chosen. Extensive comparisons with the SAX and 1d-SAX representations are included in the form of performance profiles, showing that ABBA is able to better preserve the essential shape information of time series compared to other approaches. \rev{Advantages and applications of ABBA are discussed, including its in-built differencing property and use for anomaly detection, and Python implementations provided.} 
\keywords{time series \and symbolic aggregation \and dimension reduction \and Brownian bridge}
\end{abstract}

\section{Introduction}
\label{sec:intro}

Symbolic representations of time series are an active area of research, being useful for many data mining tasks including dimension reduction, motif and rule discovery, prediction, and clustering of time series. Symbolic time series representations  allow for the use of  algorithms from text processing and bioinformatics, which often take advantage of the discrete nature of the data.  Our focus in this work is to develop a symbolic representation which is dimension reducing whilst preserving the essential \emph{shape} of the time series. Our definition of shape is  different from the one commonly implied in the context of time series: we focus on representing the peaks and troughs of the time series in their correct order of appearance, but we are happy to slightly stretch the time series in \emph{both} the time and value directions. In other words, our focus is not necessarily on approximating the time series values at the correct time points, but on representing the local up-and-down behavior of the time series and identifying repeated motifs. This is obviously not appropriate in all applications, but we believe it is close to how humans summarize the overall \rev{behavior} of a time series, and in that our representation might be useful for trend prediction, anomaly detection, and motif discovery. 

\begin{figure}[h] 
\centering 
\includegraphics[scale=1]{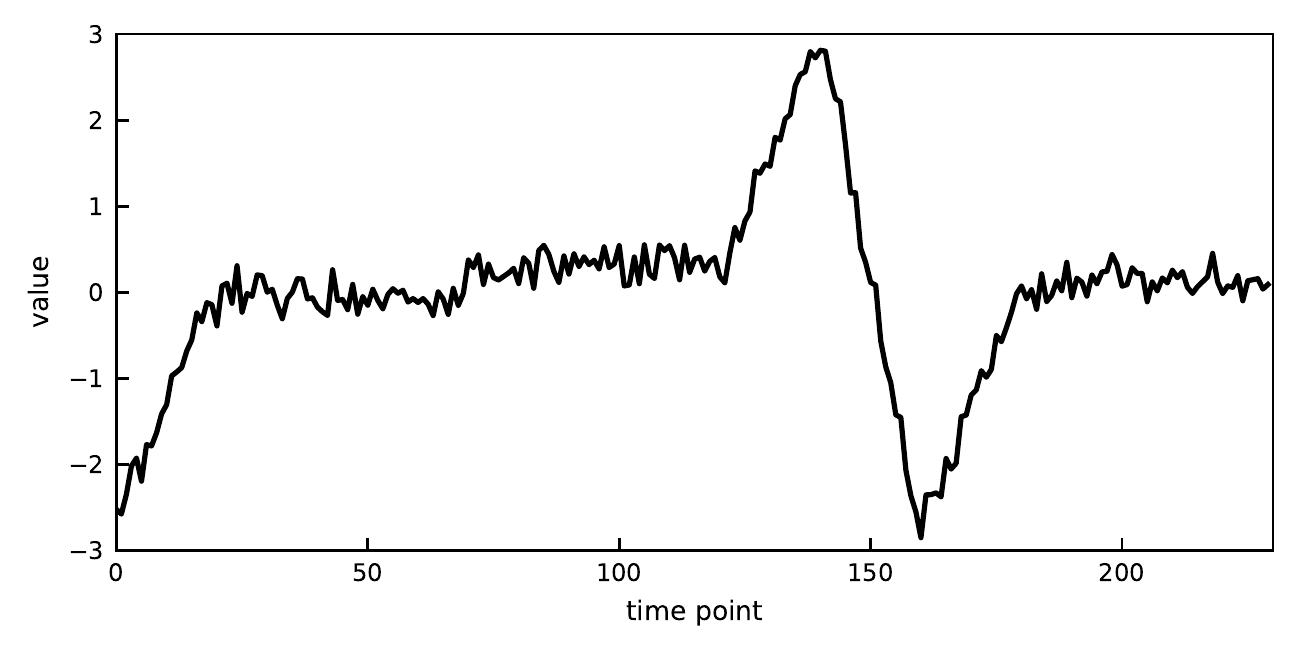}
\caption{Illustrative example time series $T$ used throughout the paper.}
\label{fig:examplets}
\end{figure}

To illustrate, let us consider the time series shown in Figure~\ref{fig:examplets}. This series is sampled at equidistant time points with values   $t_0,t_1,\ldots, t_N\in\mathbb{R}$, where $N=230$. There are various ways of describing this time series, for example:
\begin{itemize}
\item[(a)] It is exactly representable as a high-dimensional vector $T = [ t_0, t_1, \ldots, t_N] \in \mathbb{R}^{N+1}$.
\item[(b)] It starts at a value of about $-3$, then climbs up to a value of about $0$ within 25 time steps, then it stays at about~$0$  for 100 time steps, after which it goes up to a value of about $3$ within 25 time steps, and so on.
\item[(c)] It starts at a value of about $-3$, then goes up rapidly by about $3$ units, followed by a longer period with almost no change in value, after which it \emph{again} goes up rapidly by about $3$ units, and so on.
\end{itemize}

Note how in (a) and (b) the emphasis is on the actual values of the time series, whereas in (c) we mainly refer to trends in the time series in relation to previously observed trends. High-level information might be difficult to extract from (a) directly, while (b) could be seen as putting too much emphasis on the time series values instead of the overall shape. The symbolic representation developed in this paper, called \emph{adaptive Brownian bridge-based aggregation} (ABBA), adaptively reduces~$T$ to a shorter sequence of symbols with an emphasis on the  shape information. The resulting description will be conceptually  similar to (c) from the examples above. 

To formalize the discussion and introduce notation, we consider the problem of aggregating a time series $T = [ t_0, t_1, \ldots, t_N] \in \mathbb{R}^{N+1}$ into a \emph{symbolic representation} $S = [ s_1, s_2, \ldots, s_n ]\in \mathbb{A}^n$, where each $s_j$ is an element of an alphabet $\mathbb{A} = \{ a_1, a_2, \ldots, a_k\}$ of $k$ symbols. The sequence $S$ should be of considerably lower dimension than the original time series~$T$, that is $n \ll N$, and it should only use a small number of meaningful symbols, that is $k \ll n$. The representation should also allow for the approximate reconstruction of the original time series with a controllable error, with the shape of the reconstruction suitably close to that of the original. Both $n$, the length of the symbolic representation, and $k$, the number of symbols, should be chosen automatically without parameter tuning required.  

This paper is organized as follows. In Section~\ref{sec:background} we give an overview of existing symbolic representations and other algorithms which are conceptually similar to ABBA. To evaluate the approximation accuracy of ABBA, we must compare the shape of the original time series and the  reconstruction from its symbolic representation. Section~\ref{sec:distance} reviews existing distance measures for this purpose and discusses how well they perform in measuring shape. \rev{Sections~\ref{sec:method}--\ref{sec:appl} contain the key contributions of this paper:
\begin{itemize}
\item Section~\ref{sec:method} introduces ABBA, our novel dimension-reducing symbolic time series representation which aims to preserve the shape of the original time series. We explain in detail how ABBA's compression and reconstruction procedures work. 
\item In Section~\ref{brownian_bridge} we show that the error of the ABBA reconstruction behaves like a random walk with pinned start and end values. This observation appears to be novel in itself and allows us to balance the error of the piecewise linear approximation with that of the digitization procedure, thereby allowing the method to choose the number of symbols~$k$ automatically.
\item Section~\ref{sec:experiment} contains performance comparisons of ABBA with other popular symbolic representations using various distance measures, with a particular emphasis on the compression versus accuracy relation. Aside from verifying that ABBA can represent time series to higher accuracy than SAX and 1d-SAX using a comparable number of symbols $k$ and string length~$n$, we also find that SAX outperforms 1d-SAX when the same number of symbols~$k$ is used for both. 
\item In Section~\ref{sec:appl} we discuss some practical applications of ABBA including the handling of linear trends, anomaly detection, and VizTree visualization.
\end{itemize}
Finally, we conclude in Section~\ref{sec:conc} with an outlook on future work.
}

\section{Background and related work}\label{sec:background}

Despite the large number of dimension-reducing time series representations in the literature, very few are \emph{symbolic}. Most techniques are \emph{numeric} in the sense that they reduce a time series to a lower-dimensional vector with its components taken from a continuous range; see \cite{BR14,FU11,LKWL07} for reviews. Here we provide an overview of existing symbolic representations relevant to ABBA. 

The construction of symbolic time series representations typically consists of two parts. First, the time series is segmented, with the length of each segment being either specified by the user or found adaptively via a bottom-up, top-down, or sliding window approach \cite{KCHP01}. The segmentation procedure intrinsically controls the degree of dimension reduction. The second part, the discretization process, assigns a symbol to each segment.

\rev{\emph{Symbolic Aggregate approXimation}} (SAX), a very popular symbolic representation, consists of a piecewise approximation of the time series followed by a symbolic conversion using Gaussian breakpoints \cite{LKWL07}. SAX starts by partitioning $T$ into segments of constant  length $\texttt{len}$, and then represents each segment by the mean of its values (i.e., a piecewise constant approximation). The means are converted into symbols using breakpoints that partition a Gaussian bell curve into $k$ equally-sized areas. In addition to its simplicity, an attractive feature of SAX is the existence of distance measures that serve as lower bounds for the Euclidean distance between the original time series. On the other hand, both the segment length $\texttt{len}$ and the number of symbols $k$ must be specified in advance. SAX is designed such that each symbol appears with equal probability, which works best when the time series values are approximately normally distributed.

The literature on applications of SAX is extensive and many variants have been proposed. Most variants modify the symbolic representation to incorporate the slope of the time series on each segment. This is often justified by applications in finance, where the extreme values of time series provide valuable information which is lost with the piecewise constant approximation used in SAX. The modifications \rev{often} come at the cost of losing the lower bounds on distance measures. We now provide a brief overview of some of these variants.

\emph{Trend-based and Valued-based Approximation} (TVA) uses SAX to symbolically represent the time series values, enhanced with U, D, or S symbols to represent an upwards, downwards, or straight trend, respectively \cite{EART12}. The TVA representation alternates between value symbols and slope symbols, making the symbolic representation twice as long as a SAX representation with the same number of segments. A similar approach is \emph{Trend-based SAX} (TSAX) which \rev{uses two trend symbols per} segment \cite{ZLCH18}.

\emph{Extended SAX} (ESAX) represents each segment by the minimum, maximum, and mean value of the time series ordered according to their appearance in the segment, defining the mean to appear in the center of the segment~\cite{LSK06}. This results in a symbolic representation three times longer  than the corresponding SAX representation with the same number of segments. \rev{\emph{ENhanced SAX} (EN-SAX) forms a vector for each segment consisting of the minimum, maximum and mean value. The vectors are then clustered and a symbol is allocated to each cluster \cite{BBO12}.} \rev{\emph{Time-Weighted Average for SAX} (TWA\_SAX) uses the time weighted average for each segment instead of the mean \cite{BBH15}. This can encapsulate important patterns which are missed by the mean.}

\emph{Trend-based Symbolic approximation} (TSX) represents each segment by four symbols \cite{LZY12}. The first symbol corresponds to the SAX representation. The following three symbols correspond to the slopes between the first, last, most peak and most dip points, which are defined in terms of vertical distance from the trend line (the straight line connecting the end point values of a segment). The slopes are converted to symbols using a lookup table. This results in a symbolic representation four times longer than the SAX representation with the same number of segments.

The \emph{1d-SAX algorithm} uses linear regression to fit a straight line to each segment \cite{MGQT13}. Each segment is then represented by the gradient and the average value of the line. Two sets of Gaussian breakpoints are used to provide symbols for both the averages and the slopes. It is unclear how many breakpoints should be allocated for the averages, and how many should be allocated for the slopes. The total number of symbols is the product of the respective number of breakpoints. 

Using the same number of segments, the above SAX variants result in an increase in the length of the symbolic representation by some factor. It is unclear whether any of these approaches performs better than SAX when the SAX segment length $\texttt{len}$ is decreased by the same factor (keeping the overall length of the symbolic representation constant). As with the original SAX approach, all of these variants require the user to specify the segment length~$\texttt{len}$ and the number of symbols~$k$ in advance.

In many time series applications, the assumption that the values of the normalized time series follow a normal distribution is a strong one. To overcome this, the \emph{adaptive SAX algorithm} (aSAX) uses $k$-means clustering to find the breakpoints for the symbolic conversion \cite{PLD10}. However, as piecewise constant approximations are used, the aSAX approach fails to represent the extreme points of the time series. 

\rev{SAX's digitization procedure based on Gaussian breakpoints allows its extension to a multi-resolution symbolic representation known as indexable SAX (iSAX) \cite{SK08}. This clever indexing procedure allows mining of datasets containing millions of time series. At the heart of the algorithm is a SAX representation where each window uses Gaussian breakpoints with $2^c$ regions, where $c$ can change from segment to segment.}

The \emph{sensorPCA algorithm}  overcomes the fixed window length problem by using a sliding window to start a new segment when the standard deviation of the approximation exceeds some prespecified tolerance \cite{GBC13}. However, \cite{GBC13} does not provide a method to convert the mean values and window lengths to a symbolic representation.

\rev{\emph{Symbolic Aggregate approXimation Optimized by data}} (SAXO)  is a data-driven approach based on a regularized Bayesian coclustering method called  minimum optimized description length \cite{BBC16,B01}. The discretization of the time series is optimized using Bayesian statistics. The number of symbols and the underlying distribution change for each time interval. The computational complexity of SAXO is far greater than that of SAX.

The authors in \cite{MU06} take a completely different approach based on the persistence of a time series. A persistent time series is one where the value at a certain point is closely related to the previous value; see also \cite{K00}. The authors provide ``persist'', a symbolic representation based on the Kullback--Leibler divergence between the marginal and the self-transition probability distributions of the discretization symbols. 

\rev{Symbolic Polynomial (SP) \cite{GWS14} is a symbolic representation designed to detect local patterns. It is constructed by an overlapping sliding window of length~$w$ and stepsize $1$. For each window, one computes the coefficients of a regression polynomial of degree $d$. The coefficients of each order are collected and allocated a symbol using an equi-area discretization. This symbolic representation provides no dimensional reduction as each window is represented by $d$ symbols.}

\rev{The authors in \cite{BR15} introduce a symbolic representation of multivariate time series called SMTS. They construct a data table consisting of time index, time values, and first differences of the time series. A tree learner is trained on the data and each of the leaf nodes is allocated a symbol. Their approach allows multiple tree learners, which in the univariate case results in a symbolic representation much larger than the original.}

Piecewise linear approximations of time series have been used for many years. The lengths of the linear pieces (segments) can be prespecified or chosen adaptively. Each segment is approximated using either linear interpolation or linear regression \cite{KCHP01}. The authors of \cite{LYCLFHM15} describe how the linear segments can be stitched so that each piece is represented by two parameters rather than three. An example of a piecewise linear approximation algorithm is the Ramer--Douglas--Peucker algorithm, an iterative endpoint fitting procedure which uses adaptive linear interpolation with a prespecified tolerance. These methods provide an effective shape-preserving and dimension-reducing representation but not a symbolic representation.

\section{Distance measures}\label{sec:distance}

The accuracy of a symbolic time series representation $S$ can be assessed by the distance between the original time series $T$ and its reconstruction $\widehat T$ from $S$. 
We note that the original time series should first be normalized to have zero mean and unit variance. This ensures that distance measures are comparable across different time series; see \cite{KK03} for a discussion of the importance of normalization.

\begin{figure}
\centering
\subfloat[These time series have essentially the same shape but there is a value shift on the intervals {$[20, 40]$ and $[60, 80]$}.]{%
  \hspace*{-5mm}\includegraphics[scale=1]{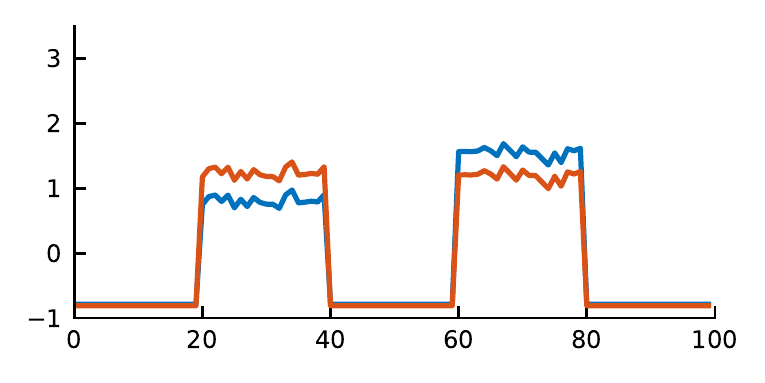}\label{fig:shift_in_value}
  }
  \hspace{0.3in}
\subfloat[These time series have essentially the same shape but they are warped in the time direction.]{%
 \hspace*{-5mm}\includegraphics[scale=1]{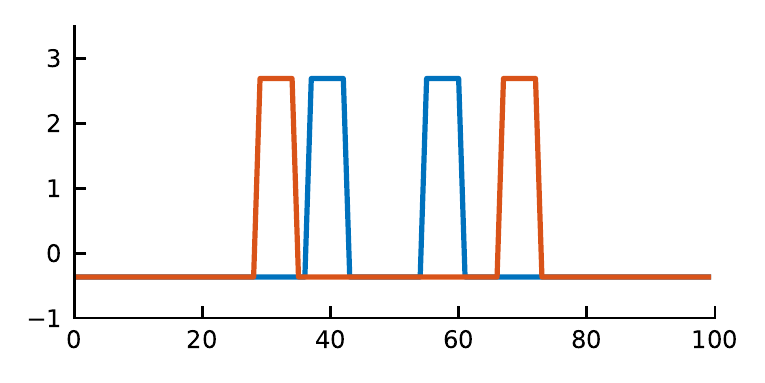}\label{fig:shift_in_time}%
  }
\caption{The time series in these plots have the same essential shape according to our interpretation. Euclidean distance is a poor measure of shape for (a) and (b), whereas DTW distance is a poor measure of shape for (a). A~differencing of the time series in (a) would make DTW a suitable shape distance.}
\label{fig:measure_flaws}
\end{figure}

A detailed overview of time series distance measures and their applications can be found in \cite{ASW15}. Distance measures for time series typically fall into two main categories: \emph{lock-step alignment} and \emph{elastic alignment} \cite{AML18}. Lock-step alignment refers to the element-wise  comparison of time series, i.e., the $i$-th element of one time series is compared to the $i$-th element of another. Such measures can only compare time series of equal length. The most popular lock-step distance is the Euclidean distance. 
The Euclidean distance is a poor measure of shape similarity in two particular cases: if the time series have the same shape but are stretched in value (see Figure~\ref{fig:shift_in_value}), or if the time series have the same shape but are warped in time (see Figure~\ref{fig:shift_in_time}). The first issue can be mitigated by differencing the time series before measuring the distance. The second issue is intrinsic to lock-step alignment distance measures.

Elastic alignment distance measures construct a nonlinear mapping between time series elements, effectively allowing for  one value in a time series to be compared to multiple consecutive values in another. The most popular elastic alignment method is Dynamic Time Warping (DTW), originally proposed in \cite{BC94}. The DTW distance measure corresponds to the Euclidean distance between two DTW-aligned time series. This distance measure can be used to compare time series of different lengths but it has a quadratic computational complexity in both time and space; for further details see \cite{KR05}. Many methods have been proposed to either approximate the DTW distance at a reduced cost or calculate bounds to avoid computing the DTW alignment  altogether. The authors of \cite{KP01} notice that DTW may pair a rising trend in one time series with a falling trend in another, and they overcome this problem by a variant known as Derivative Dynamic Time Warping (DDTW). The elastic alignment allows DTW to overcome the issues when two time series have the same shape but are warped in time (see Figure~\ref{fig:shift_in_time}), but DTW is still a poor measure of shape similarity if the time series have the same shape but are vertically stretched (see Figure~\ref{fig:shift_in_value}). Again, this can be mitigated by differencing the time series before measuring their DTW distance. 

It is because of these advantages and drawbacks of the Euclidean and DTW distance measures and their differenced counterparts that we will test the performance of ABBA with all these distance measures in Section~\ref{sec:experiment}.

\section{Adaptive \rev{Brownian bridge-based aggregation}} \label{sec:method}
We now introduce ABBA, a symbolic representation of time series where the symbolic length $n$ and the number of symbols $k$ are chosen adaptively. 
The ABBA representation is computed in two stages.
\begin{enumerate}
\item Compression: The original time series $T$ is approximated by a piecewise linear and continuous function, with each linear piece being chosen adaptively based on a user-specified tolerance. The result is a sequence of tuples $(\texttt{len}, \texttt{inc})$ consisting of the length of each piece and its increment in value.
\item Digitization: A near-optimal alphabet $\mathbb{A}$ is identified via mean-based clustering, with each cluster corresponding to a symbol. Each tuple $(\texttt{len}, \texttt{inc})$ is assigned a symbol corresponding to the cluster in which it belongs.
\end{enumerate}

The reconstruction of a time series from its ABBA representation involves three stages.
\begin{enumerate}
\item Inverse-digitization: Each symbol of the symbolic representation is replaced with the center of the associated cluster. The length values of the centers may not necessarily be integers. 
\item Quantization: The lengths of the reconstructed segments are  re-aligned with an integer grid.
\item Inverse-compression: The piecewise linear continuous approximation is converted back to a pointwise time series representation using a stitching procedure.
\end{enumerate}

Both the computation of the ABBA representation and the reconstruction are inexpensive. It is essential that the digitization process uses incremental changes in value rather than slopes. This way, ABBA consistently works with increments in both the time and value coordinates. Only in this case a mean-based clustering algorithm will identify meaningful clusters in both coordinate directions. As we will explain in Section~\ref{brownian_bridge}, the error of the ABBA reconstruction behaves like a random walk pinned at zero for both the start and the end point of the time series. But first, we provide a more detailed explanation of the key parts of ABBA. For clarity, we summarize the notation used throughout this section in Table~\ref{tab:notation}.

\begin{table}[h!]
\caption {Summary of notation} \label{tab:notation} 
\centering
\renewcommand*{\arraystretch}{1.5}
\begin{tabular}{|l|l|}
\hline
Original time series: & $T = [t_0, t_1, \ldots, t_N] \in \mathbb{R}^{N+1}$ \\
\hline
After compression: & $[(\texttt{len}_1, \texttt{inc}_1), (\texttt{len}_2, \texttt{inc}_2), \ldots, (\texttt{len}_n, \texttt{inc}_n)] \in \mathbb{R}^{2 \times n}$ \\
\hline
After digitization: & $S=[s_1, s_2, \ldots, s_n] \in \mathbb{A}^n$ with $\mathbb{A} = \{ a_1, a_2, \ldots, a_k \}$ \\
\hline
After inverse-digitization: & $[(\widetilde{\texttt{len}}_1, \widetilde{\texttt{inc}}_1), (\widetilde{\texttt{len}}_2, \widetilde{\texttt{inc}}_2), \ldots, (\widetilde{\texttt{len}}_n, \widetilde{\texttt{inc}}_n)] \in \mathbb{R}^{2 \times n}$ \\
\hline
After quantization: & $[(\widehat{\texttt{len}}_1, \widehat{\texttt{inc}}_1), (\widehat{\texttt{len}}_2, \widehat{\texttt{inc}}_2), \ldots, (\widehat{\texttt{len}}_n, \widehat{\texttt{inc}}_n)] \in \mathbb{R}^{2 \times n}$ \\
\hline 
After inverse-compression: & $\widehat{T} = [\widehat{t}_0, \widehat{t}_1, \ldots, \widehat{t}_N] \in \mathbb{R}^{N+1}$\\
\hline
\end{tabular}
\end{table}

\subsection{Compression}
The ABBA compression is achieved by an adaptive piecewise linear continuous approximation of $T$. Given  a tolerance $\texttt{tol}$, the method adaptively selects $n+1$ indices 
$i_0 = 0 < i_1 <\cdots < i_n = N$ so that the time series $T = [t_0,t_1,\ldots, t_N]$ is approximated  by a polygonal chain going through the points $(i_j , t_{i_j})$ for $j=0,1,\ldots,n$. This gives rise to a partition of $T$ into $n$ pieces $P_j = [ t_{i_{j-1}},t_{i_{j-1}+1},\ldots, t_{i_j} ]$, each of length $\texttt{len}_j := i_j - i_{j-1}\geq 1$ in the time direction. 
We ensure that the squared Euclidean distance of the values in $P_j$ from the straight polygonal line is bounded by $(\texttt{len}_j - 1)\cdot\texttt{tol}^2$. 
More precisely, starting with $i_0 = 0$ and given an index $i_{j-1}$, we find the largest possible $i_j$ such that $i_{j-1} < i_j\leq N$ and 
\begin{equation}
\sum_{i=i_{j-1}}^{i_j} \left( \underbrace{t_{i_{j-1}} + (t_{i_j} - t_{i_{j-1}})\cdot \frac{i - i_{j-1}}{i_j - i_{j-1}}}_{\text{straight line approximation}} \ \ - \underbrace{t_i}_{\text{actual value}} \right)^2 \leq (i_{j} - i_{j-1} -1)\cdot\texttt{tol}^2.
	\label{eq:compress}
\end{equation}
Note that the first and the last values $t_{i_{j-1}}$ and $t_{i_{j}}$ are not counted in  the distance measure as the straight line approximation passes exactly through them. 
If required, one can restrict the maximum length  of each segment by imposing an upper bound $i_{j} \leq i_{j-1} + \texttt{max\_len}$ with a given integer $\texttt{max\_len}\geq 1$.

Each linear piece $P_j$ of the resulting polygonal chain $\widetilde T$ is described by a tuple $(\texttt{len}_j, \texttt{inc}_j)$, where $\texttt{inc}_j = t_{i_j} - t_{i_{j-1}}$ is the increment in value (not the slope!). As the polygonal chain is continuous, the first value of a segment can be inferred from the end value of the previous segment. Hence the whole polygonal chain can be recovered exactly from the first value $t_0$ and the tuple sequence 
\begin{equation}
(\texttt{len}_1, \texttt{inc}_1), (\texttt{len}_2, \texttt{inc}_2), \ldots, (\texttt{len}_n, \texttt{inc}_n)\in \mathbb{R}^2.
\label{eq:tupseq}
\end{equation}

\begin{figure}[h] 
\centering 
\begin{minipage}[t]{0.47\textwidth}
	\hspace*{-3mm}    
	\includegraphics[scale=1]{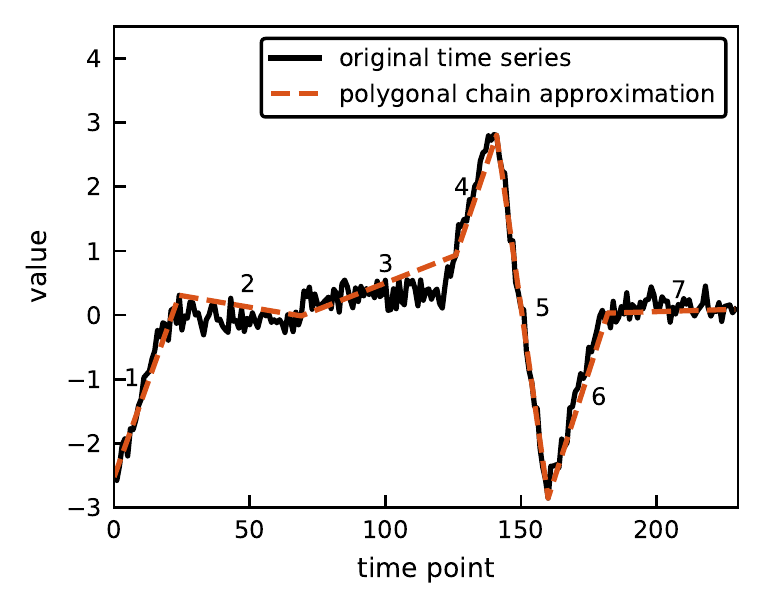} 
    \caption{Result of the ABBA compression. The time series is now represented by $n=7$ tuples of the form $(\texttt{inc}, \texttt{len})$ and the starting value $t_0$.}
    \label{fig:compression}
\end{minipage}\hfill
\begin{minipage}[t]{0.47\textwidth}
    \hspace*{-3mm} 
    \includegraphics[scale=1.018]{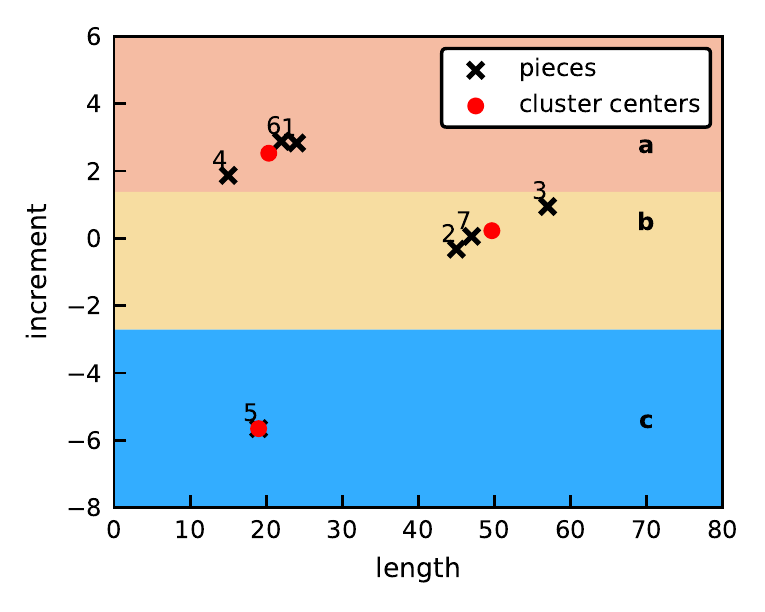} 
    \caption{Result of the ABBA digitization with scaling parameter $\texttt{scl} = 0$. The tuples $(\texttt{len}, \texttt{inc})$ are converted to the symbol sequence $\texttt{abbacab}$.}
    \label{fig:digitization0}
\end{minipage}
\end{figure}

An example of the ABBA compression procedure applied to the time series in Figure~\ref{fig:examplets} is shown in Figure~\ref{fig:compression}. Here a tolerance of $\texttt{tol} = 0.4$ has been used, resulting in $n=7$ pieces. As the approximation error on each piece~$P_j$ satisfies \eqref{eq:compress}, the polygonal chain $\widetilde{T}$ also has a bounded Euclidean distance from $T$:  
\begin{align}
\texttt{euclid}(T, \widetilde{T})^2 &\leq  [(i_1 - i_0 -1) + (i_2 - i_1 - 1) + \cdots + (i_n - i_{n-1} - 1) ]\cdot\texttt{tol}^2\label{eq:euclid} \\
&= (N  -n)\cdot \texttt{tol}^2.\nonumber
\end{align}
Hence we are sure that the ABBA approximation $\widetilde{T}$ (red dashed curve) in Figure~\ref{fig:compression} has a Euclidean distance of at most $\sqrt{223}\times 0.4\approx 6.0$ from the original time series $T$ (black solid curve).

\subsection{Digitization}\label{sec:digit}
Digitization refers to the assignment of the tuples in \eqref{eq:tupseq} to $k$ clusters $S_1,S_2,\ldots,S_k$.  
Before clustering, we separately normalize the tuple lengths and increments by their standard deviations $\sigma_{\texttt{len}}$ and $\sigma_{\texttt{inc}}$, respectively. We use a further scaling parameter $\texttt{scl}$ to assign different weight (``importance'') to the length of each piece in relation to its increment value. Hence, we effectively cluster the \emph{scaled tuples} 
\begin{equation}
\left(\texttt{scl}\frac{\texttt{len}_1}{\sigma_{\texttt{len}}}, \frac{\texttt{inc}_1}{\sigma_{\texttt{inc}}}\right), \left(\texttt{scl}\frac{\texttt{len}_2}{\sigma_{\texttt{len}}}, \frac{\texttt{inc}_2}{\sigma_{\texttt{inc}}} \right), \ldots, \left(\texttt{scl}\frac{\texttt{len}_n}{\sigma_{\texttt{len}}}, \frac{\texttt{inc}_n}{\sigma_{\texttt{inc}}} \right) \in \mathbb{R}^2.
\label{eq:tupseq2}
\end{equation}
If $\texttt{scl} = 0$, then clustering is performed on the increments alone, while if $\texttt{scl} = 1$, we cluster in both the length and increment dimension with equal weighting. The cluster assignment is performed by (approximately) minimizing the within-cluster-sum-of-squares 
\begin{equation}
\texttt{WCSS} = \sum_{i=1}^{k} \sum_{(\texttt{len}, \texttt{inc}) \in S_i} \Bigg\Vert \left(\texttt{scl}\frac{\texttt{len}}{\sigma_{\texttt{len}}}, \frac{\texttt{inc}}{\sigma_{\texttt{inc}}}\right) - \overline{\mu}_i \Bigg\Vert^2 ,
\nonumber 
\end{equation}
with each 2d cluster center $\overline{\mu}_i = (\overline{\mu}_i^{\texttt{len}},\overline{\mu}_i^{\texttt{inc}})$ corresponding to the mean of the {scaled} tuples associated with the cluster $S_i$. 
In certain situations one may want to cluster only on the lengths of the pieces and ignore their increments, formally setting $\texttt{scl} = \infty$. In this case, the cluster assignment is performed by (approximately) minimizing 
\[
\texttt{WCSS} = \sum_{i=1}^{k} \sum_{(\texttt{len}, \texttt{inc}) \in S_i} \Bigg| \frac{\texttt{len}}{\sigma_{\texttt{len}}} - \overline{\mu}_i^{\texttt{len}} \Bigg|^2 ,
\]
where $\overline{\mu}_i^{\texttt{len}}$ is the mean of the scaled lengths in the cluster $S_i$.

Given a clustering of the $n$ tuples into clusters $S_1,\ldots,S_k$ we use the \emph{unscaled} cluster centers $\mu_i$
\begin{equation}
 \mu_i = (\mu_i^{\texttt{len}}, \mu_i^{\texttt{inc}}) = \frac{1}{|S_i|} \sum_{(\texttt{len}, \texttt{inc}) \in S_i} (\texttt{len}, \texttt{inc})
 \nonumber
\end{equation}
to define the maximal cluster variances in the length and increment directions as 
\begin{eqnarray*}
\label{eq:lenvar}
\mathrm{Var}_{\texttt{len}} &=&  \max_{i=1,\ldots,k} \frac{1}{|S_i|} \sum_{(\texttt{len}, \texttt{inc}) \in S_i} \left|   \texttt{len} - \mu_i^{\texttt{len}}\right|^2, \\
\mathrm{Var}_{\texttt{inc}} &=&  \max_{i=1,\ldots,k} \frac{1}{|S_i|} \sum_{(\texttt{len}, \texttt{inc}) \in S_i} \left|\texttt{inc} - \mu_i^{\texttt{inc}}\right|^2,
\label{eq:incvar}
\end{eqnarray*}
respectively.  Here, $|S_i|$ is the number of tuples in cluster $S_i$.  
We seek the smallest number of clusters $k$ such that 
\begin{equation}\label{eq:tolbnd}
\max(\texttt{scl}\cdot\mathrm{Var}_{\texttt{len}}, \mathrm{Var}_{\texttt{inc}}) \leq \texttt{tol}_s^2
\end{equation}
with a tolerance $\texttt{tol}_s$. This tolerance will be specified in Section~\ref{brownian_bridge} as a function of the user-specified tolerance $\texttt{tol}$ and is therefore not a free parameter. (In the case of $\texttt{scl} = \infty$, we seek the smallest $k$ such that $
\mathrm{Var}_{\texttt{len}} \leq  \texttt{tol}_s^2$.) Once the optimal $k$ has been found, each cluster $S_1, \ldots, S_k$ is assigned a symbol $a_1,  \ldots, a_k$, respectively. Finally, each tuple in the sequence \eqref{eq:tupseq} is replaced by the symbol of the cluster it belongs to, resulting in the symbolic representation $S = [ s_1,s_2,\ldots,s_n]$.

If $\texttt{scl} = 0$ or $\texttt{scl} = \infty$, a 1d clustering method can be used which takes advantage of sorting algorithms; see the review  \cite{GLMN17}. We use the \texttt{ckmeans} algorithm \cite{WS11}, an order $\mathcal{O}(n \log n+ kn)$ dynamic programming algorithm which optimally clusters the data by minimizing the WCSS in just one dimension. We have modified the algorithm to choose the smallest~$k$ such that the maximal cluster variance is bounded by $\texttt{tol}_s^2$.

For nonzero finite values of $\texttt{scl}$, $k$-means clustering is used. This algorithm  has an average complexity of $\mathcal{O}(kn)$ per iteration (see also \cite{AV06} for an analysis of the worst case complexity) and might of course result in a suboptimal clustering. In our ABBA implementation the user can specify an interval $[\texttt{min\_k}, \ldots, \texttt{max\_k}]$ and we search for the smallest $k$ in that interval such that \eqref{eq:tolbnd} holds.  If no such $k$ exists,  we set $k = \texttt{max\_k}$. 

By default, we set $\texttt{scl}=0$ as we believe this corresponds most naturally to preserving the up-and-down behavior of the time series. In other words, we ignore the lengths of the pieces and only cluster the value increments. With the value increments represented accurately, the errors in lengths correspond to horizontal stretching in the time direction.

\begin{figure}[h] 
\centering 
\begin{minipage}[t]{0.47\textwidth}
    \hspace*{-3mm}\includegraphics[scale=1]{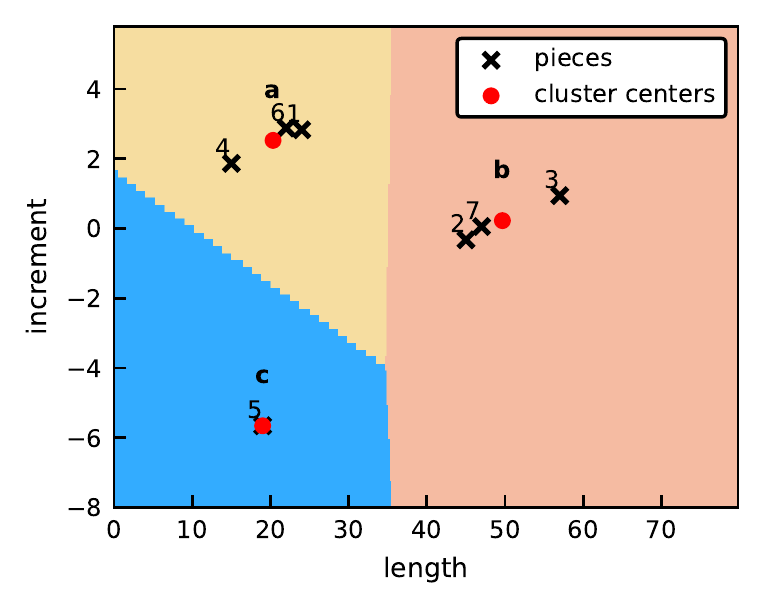} 
    \caption{Result of the ABBA digitization with $\texttt{scl} = 1$. The tuples $(\texttt{len}, \texttt{inc})$ are converted to the symbol  sequence  $\texttt{abbacab}$.}
    \label{fig:digitization1}
\end{minipage}\hfill
\begin{minipage}[t]{0.47\textwidth}
    \hspace*{-3mm}
    \includegraphics[scale=1.018]{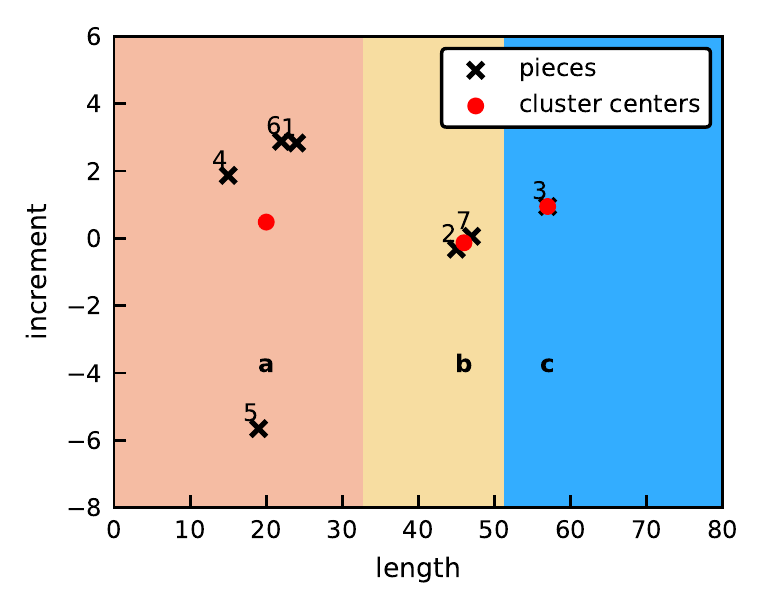} 
    \caption{Result of the ABBA digitization with $\texttt{scl} = \infty$. The tuples $(\texttt{len}, \texttt{inc})$ are converted to the symbol  sequence  $\texttt{abcaaab}$.}
    \label{fig:digitizationinf}
\end{minipage}
\end{figure}

An illustration of the digitization process on the pieces from Figure~\ref{fig:compression} can be seen in Figure~\ref{fig:digitization0} with $\texttt{scl} = 0$ (our default parameter choice), Figure~\ref{fig:digitization1} with $\texttt{scl} = 1$, and Figure~\ref{fig:digitizationinf} with $\texttt{scl} = \infty$.

\subsection{Inverse digitization and quantization}

When reversing the digitization process, each symbol of the alphabet is  replaced by the center $(\overline{\texttt{len}}_i, \overline{\texttt{inc}}_i)$ of the corresponding cluster given as
\begin{equation}
 (\overline{\texttt{len}}_i, \overline{\texttt{inc}}_i) = \frac{1}{|S_i|} \sum_{(\texttt{len}, \texttt{inc}) \in S_i} (\texttt{len}, \texttt{inc}).
\nonumber 
\end{equation}
Note that the mean-based clustering for digitization is performed on the scaled tuples \eqref{eq:tupseq2}, but the cluster centers used for the inverse digitization are computed with the unscaled tuples \eqref{eq:tupseq}.
The inverse digitization process results in a sequence of $n$ tuples 
$$
(\widetilde{\texttt{len}}_1, \widetilde{\texttt{inc}}_1), (\widetilde{\texttt{len}}_2, \widetilde{\texttt{inc}}_2), \ldots, (\widetilde{\texttt{len}}_n, \widetilde{\texttt{inc}}_n) \in \mathbb{R}^{2},
$$
where each tuple is a cluster center, that is $(\widetilde{\texttt{len}}_i, \widetilde{\texttt{inc}}_i) \in \{(\overline{\texttt{len}}_1, \overline{\texttt{inc}}_1), (\overline{\texttt{len}}_2, \overline{\texttt{inc}}_2), \ldots, (\overline{\texttt{len}}_k, \overline{\texttt{inc}}_k)\}$. 

The lengths $\widetilde{\texttt{len}}_i$ obtained from this averaging are not necessarily integer values as they were in the compressed representation~\eqref{eq:tupseq}. 
We therefore perform a simple quantization procedure which realigns the cumulated lengths with their closest integers. We start with rounding the first length, $\widehat{\texttt{len}}_1 := \mathrm{round}(\widetilde{\texttt{len}}_1)$, keeping track of the rounding error $e := \widetilde{\texttt{len}}_1 - \widehat{\texttt{len}}_1$. This error is added to the second length $\widetilde{\texttt{len}}_2 := \widetilde{\texttt{len}}_2 + e$, which is then rounded to $\widehat{\texttt{len}}_2 := \mathrm{round}(\widetilde{\texttt{len}}_2)$ with error 
$e := \widetilde{\texttt{len}}_2 - \widehat{\texttt{len}}_2$, and so on. As a result we obtain a sequence of $n$ tuples
\begin{equation}
(\widehat{\texttt{len}}_1, \widehat{\texttt{inc}}_1), (\widehat{\texttt{len}}_2, \widehat{\texttt{inc}}_2), \ldots, (\widehat{\texttt{len}}_n, \widehat{\texttt{inc}}_n) \in \mathbb{R}^{2}
\label{eq:quant}
\end{equation}
with integer lengths $\widehat{\texttt{len}}_i$. (The increments remain unchanged but we rename them for consistency: $\widehat{\texttt{inc}}_i := \widetilde{\texttt{inc}}_i$.)

\section{Error analysis}\label{brownian_bridge}

During the compression procedure, we construct a polygonal chain $\widetilde T$ going through selected points $\{ (i_j, t_{i_j}) \}_{j = 0}^{n}$ of the original time series $T$, with a controllable Euclidean distance \eqref{eq:euclid}. After the digitization, inverse digitization, and quantization, we obtain a new tuple sequence \eqref{eq:quant} which can be stitched together to a polygonal chain $\widehat T$ going through the points $\{ (\widehat{i}_j, \widehat{t}_{j}) \}_{j = 0}^{n}$, with $(\widehat{i}_0,\widehat{t}_0)=(0,t_0)$. Our aim is to analyze the distance between $\widehat T$ and $\widetilde T$, and then balance it with the distance between $\widetilde T$ and $T$. 

We first note that 
\[
(\widehat{i}_j, \widehat{t}_{i_j} ) = \left(\sum_{\ell = 1}^{j} \widehat{\texttt{len}}_{\ell}, \: t_0 + \sum_{\ell = 1}^{j} \widehat{\texttt{inc}}_{\ell}\right), \quad j=0,\ldots,n.
\] 
As all the lengths $\widehat{\texttt{len}}_{\ell}$ and increments $\widehat{\texttt{inc}}_{\ell}$ correspond to cluster centers (averages of all the points in a cluster, consistently rounded during quantization), we have the interesting property that the accumulated deviations from the true lengths and increments exactly cancel out at the right endpoint of the last piece $P_n$, that is: $(\widehat{i}_n, \widehat{t}_{i_n} ) = (i_n, t_{i_n}) = (N, t_N)$. In other words, the polygonal chain $\widehat T$ starts and ends at the same values as $\widetilde T$ (and hence $T$).

\begin{figure}[h] 
\centering 
\begin{minipage}[t]{0.47\textwidth}
    \centering
    \hspace*{-3mm}\includegraphics[width=1.1\textwidth]{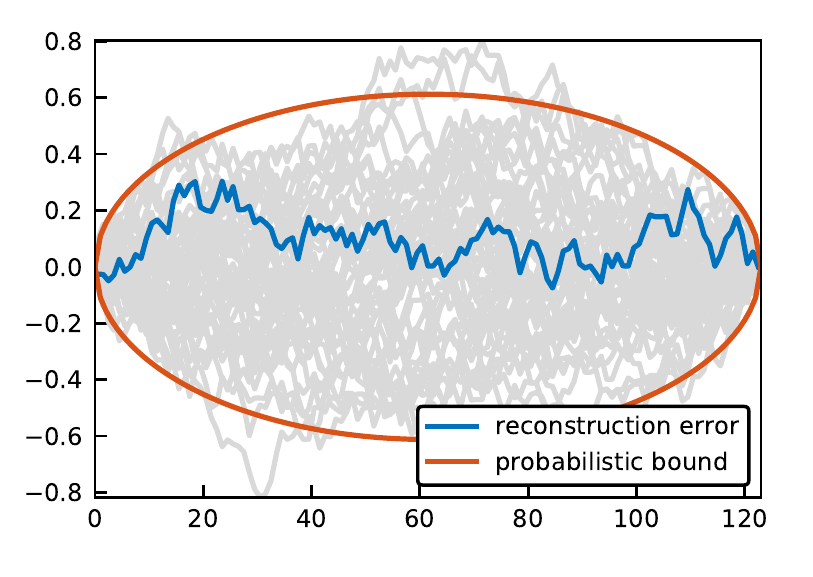} 
    \caption{Example of the ABBA reconstruction error forming a Brownian bridge. The blue line is the actual error, the grey lines are 50 other realizations of the random walk, and the red bounds indicate one standard deviation above and below the zero mean.}
    \label{fig:brownian_bridge}
\end{minipage}\hfill
\begin{minipage}[t]{0.47\textwidth}
    \centering
    \hspace*{-6mm}\includegraphics[width=1.11\textwidth]{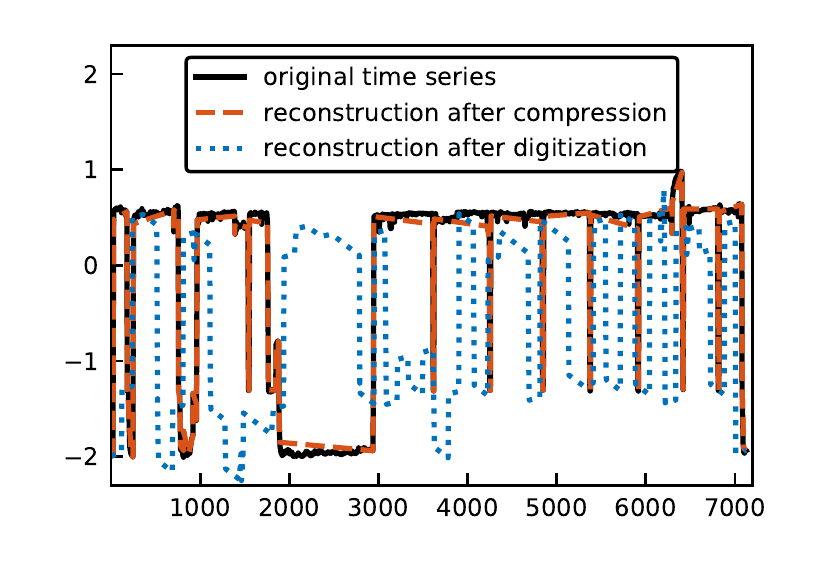} 
    \caption{ABBA representation of a time series from a heat exchanger in an ethylene cracker. With $\texttt{tol} = 0.1$ and $\texttt{scl} = 0$, the time series is reduced from $7128$ points to $123$ tuples using $14$ symbols.}
    \label{fig:decoke}
\end{minipage}
\end{figure}

We now analyze the \rev{behavior} of $\widehat T$ in between the start and endpoints, focusing on the case that $\texttt{scl}=0$ and assuming for simplicity that all cluster centers $S_i$ have the same mean length $\mu_i^\texttt{len} = N/n$. (This is not a strong assumption as in the dynamic time warping distance the lengths of the pieces is irrelevant.)  We compare $\widehat T$ with the polygonal chain $\widetilde T$ time-warped to the same regular length grid as $\widehat T$, which will give an upper bound on $\texttt{dtw}(\widehat T,\widetilde T)$. Denoting by $d_\ell := \widehat{\texttt{inc}}_{\ell} - \widetilde{\texttt{inc}}_{\ell}$ the local deviation of the increment value of $\widehat T$ on piece $P_\ell$ from the true increment of $\widetilde T$, we have that
\[
 \widehat t_{i_j} - t_{i_j} = \sum_{\ell=1}^j d_\ell =: e_{i_j}, \quad j=0,\ldots,n. 
\]
Recall from Section~\ref{sec:digit} that we have controlled the variance of the increment values in each cluster to be bounded by $\texttt{tol}_s^2$. As a consequence, the increment deviations $d_\ell$ have bounded variance  $\texttt{tol}_s^2$, and mean zero as they correspond to deviations from their respective  cluster center. It is therefore reasonable to model the ``global increment errors''~$e_{i_j}$ as a random process with fixed values $e_{i_0} = e_{i_n} = 0$, expectation $\mathrm{E}(e_{i_j})=0$, and variance
\[
	\mathrm{Var}({e_{i_j}}) = \texttt{tol}_s^2 \cdot \frac{j (n-j)}{n}, \quad j=0,\ldots,n.
\]
In the case that the $d_\ell$ are i.i.d. normally distributed, such a process is known as a \emph{Brownian bridge}. See also Figure~\ref{fig:brownian_bridge} for an illustration. 

Note that so far we have only considered the variance of the global increment errors $e_{i_j}$ at the left and right endpoints of each piece $P_j$, but we are actually interested in analyzing the error of the reconstruction $\widehat T$ on the fine time grid. To this end, we now consider a ``worst-case'' realization of $e_{i_j}$ which stays $s$ standard deviations away from its zero mean. That is, we consider a realization 
\[
	{e_{i_j}} = s\cdot  \texttt{tol}_s \cdot \sqrt{\frac{j (n-j)}{n}}, \quad j=0,\ldots,n.
\]
By piecewise linear interpolation of these errors from the coarse time grid $i_0,i_1,\ldots,i_n$ to the fine time grid $i=0,1,\ldots,N$ (in accordance with the linear stitching procedure used in ABBA), we find that 
\[
    {e_{i}} \leq \sqrt{\frac{n}{N}}\cdot s\cdot  \texttt{tol}_s \cdot \sqrt{\frac{i (N-i)}{N}}, \quad i=0,\ldots,N,
\]
using that the interpolated quadratic function on the right-hand side is concave. 
We can now bound the squared Euclidean norm of this fine-grid ``worst-case'' realization as
\[
\sum_{i=0}^N e_i^2 \leq \frac{n\cdot s^2 \cdot  \texttt{tol}_s^2}{N^2} \cdot \sum_{i=0}^N i (N-i) 
= \frac{n\cdot s^2 \cdot  \texttt{tol}_s^2}{N^2} \cdot \frac{N^3 - N}{6} 
\leq  {n\cdot s^2 \cdot  \texttt{tol}_s^2} \cdot \frac{N}{6}.
\]
This is a probabilistic bound on squared Euclidean error caused by a ``worst-case'' realization of the Brownian bridge, and thereby a probabilistic bound on the error incurred from the digitization procedure. Equating this bound with the bound \eqref{eq:euclid} on the accuracy of  the compression, we find that we should  choose
\[
\texttt{tol}_s = \frac{\texttt{tol}}{s} \sqrt{\frac{6(N-n)}{Nn}},
\]
with the user-specified tolerance $\texttt{tol}$.  We have experimentally determined that $s=0.2$ typically gives a good balance between the compression accuracy and the number of clusters determined using this criterion.

\medskip 
 
\noindent\textbf{Example:} We now illustrate the above analysis on a challenging real-world example. Consider a time series~$T$ ($N=7127$) consisting of temperature readings off a heat exchanger in an ethylene cracker. We use $\texttt{tol} = 0.1$ to compress this time series, resulting in a polygonal chain $\widetilde T$ with $n=123$ pieces and an approximation error of $\texttt{euclid}(T,\widetilde T) = 5.3 \leq \sqrt{N-n}\cdot \texttt{tol} \approx 8.4$. See Figure~\ref{fig:decoke} for a plot of the original time series $T$ and its reconstruction $\widetilde T$ after compression. 

We then run the ABBA digitization procedure with scaling parameter $\texttt{scl}=0$, resulting in a symbolic representation~$S$ of length~$n$ using $k=14$ symbols. In Figure~\ref{fig:brownian_bridge} we show the ``global increment errors''~$e_{i_j}$ of the reconstruction $\widehat T$ on each piece $P_j$, that is, the increment deviation of $\widehat T$ from $T$ at the endpoints of $P_j$, $j=1,\ldots,n$. Note how this error is pinned at zero at $j=0$ and $j=n$, and how it resembles a random walk in between. 

The reconstruction $\widehat T$ on the fine time grid is also shown in Figure~\ref{fig:decoke}.
The reconstruction error measured in the time warping distance is $\texttt{dtw}(\widetilde T, \widehat T) = 9.5$ 
and the overall error is $\texttt{dtw}(T, \widehat T) = 10.8$, both of which are approximately of the same order as $\sqrt{N-n}\cdot\texttt{tol} \approx 8.4$. Note that the ABBA reconstruction $\widehat T$ visually deviates a lot from $T$ due to the rather high tolerance we have chosen for illustration, but nevertheless, the characteristic up-and-down behavior  of $T$ is well represented in $\widehat T$, despite the high compression rate of $123/7128 \approx 1.7\,\%$. 

\section{\rev{Discussion and performance comparison}\label{sec:experiment}}
A Python implementation of ABBA, along with codes to reproduce the figures and performance comparisons in this paper, can be found at
\begin{center}
\url{https://github.com/nla-group/ABBA}
\end{center} 
When the scaling parameter is $\texttt{scl} = 0$ or $\texttt{scl} = \infty$, our implementation calls an adaptation of the univariate $k$-means algorithm from the R package \texttt{Ckmeans.1d.dp} written in $\texttt{C++}$. We use SWIG, the open-source ``Simplified Wrapper and Interface Generator'', to call $\texttt{C++}$ functions from Python. 
If $\texttt{scl} \in ( 0, \infty)$, we use the $k$-means algorithm from the Python \texttt{sklearn} library \cite{scikit}.

\rev{ABBA uses the lengths and increments of a polygonal chain on each segment to construct its symbolic time series representation. Symbolic Polynomial \cite{GWS14} (with $d=1$) and 1d-SAX \cite{MGQT13}, on the other hand, use linear regression to fit a polynomial to a window of fixed pre-specified length. As we discussed in Section~\ref{sec:background}, Symbolic Polynomial provides no dimensional reduction and was specifically designed for time series classification problems. Most other SAX variants increase the length of the symbolic representation by enhancing the string with additional characters to capture shapes and trends. It is not clear whether these representations outperform SAX with a reduced width parameter to compensate for the increased string length. A comparison of this would be interesting but is independent of ABBA's performance and out of the scope of this paper. SMTS \cite{BR15} and aSAX \cite{PLD10} use machine learning techniques to discretize their representation. SMTS is primarily designed for multivariate time series and provides no dimensional reduction. EN-SAX \cite{BBO12} and aSAX suffer from a loss of the trend information in their compression step.

For these reasons, we focus on profiling the reconstructions errors of the ABBA, SAX \cite{LKWL07}, and 1d-SAX \cite{MGQT13} algorithms, as these are most closely related and easily comparable. Note that none of the representations were primarily designed as compression algorithms. ABBA was designed to be adaptive in both segement length and alphabet cardinality, whereas SAX and 1d-SAX have many other benefits such as being hashable \cite{CKL03}, indexable \cite{SK08}, and permitting lower bounding distance measures.} Our test set consists of all time series in the UCR Time Series Classification Archive \cite{UCRArchive} with a length of at least 100 data points. There are $128,978$ such time series from a variety of applications. Although the archive is primarily intended for benchmarking time series classification algorithms, our primary focus in this paper is on the approximation performance of the symbolic representations. Our experiment consists of converting each time series $T=[t_0,t_1,\ldots,t_N]$ into its symbolic representation $S = [s_1,\ldots,s_n]$, and then measuring the distance between the reconstruction $\widehat T=[\widehat t_0,\widehat t_1,\ldots,\widehat t_N]$ and $T$ in the (differenced) Euclidean and DTW norms, respectively. 

Recall from Section~\ref{sec:background} that both SAX and 1d-SAX require a choice for the fixed segment length. In order to provide a fair comparison, we first run the ABBA compression with an initial tolerance $\texttt{tol}=0.05$. This returns $n$, the number of  required pieces to approximate $T$ to this tolerance. If $n$ turns out to be larger than $N/5$, we successively increase the tolerance by $0.05$ and rerun until a compression rate of at least 20\,\% is achieved. If a time series cannot be compressed to at least 20\,\% even at the rather crude  tolerance of $\texttt{tol}=0.5$, we consider it as too noisy and exclude it from the test. We also exclude all time series which, after ABBA compression, result in fewer than nine pieces: this is necessary because we want to use $k=9$ symbols for all compared methods.  Table~\ref{tab:tol} shows how many of the $111,889$ remaining time series were compressed at what tolerance.  The table gives evidence that most of these time series can be compressed reasonably well while maintaining a rather high accuracy. The average compression rate is 10.3\,\%.

\begin{table}[h!]
\caption{Tolerance used for the compression and the number of time series to which it was applied} \label{tab:tol}
\centering
\renewcommand*{\arraystretch}{1.5}
\begin{tabular}{|l|c|c|c|c|c|c|c|c|c|c|c|}
\hline
tolerance $\texttt{tol}$ & 0.05 & 0.10 & 0.15 & 0.20 & 0.25 & 0.30 & 0.35 & 0.40 & 0.45 & 0.50 \\
nr of time series & 75417 &  9247 & 7786 & 5855 & 2972 & 2236 & 1910 & 1670 & 2146 & 2650\\
\hline
\end{tabular}
\end{table}

After the number of pieces $n$ has been specified for a given time series $T$, we determine the fixed segment length $\texttt{len} = \lfloor (N+1)/n \rfloor$ to be used in the SAX and 1d-SAX algorithms. We then apply SAX and 1d-SAX to the first $n\cdot \texttt{len}$ points of $T$. This guarantees that all three algorithms (SAX, 1d-SAX, and ABBA) produce a symbolic representation of with $n$ pieces. If $N+1$ is not divisible by $n$, SAX and 1d-SAX are applied to slightly shorter time series than ABBA. 
The number of symbols used for the digitization is $k=9$ for all three methods. In the case of 1d-SAX this means that three symbols are used for the mean value, and three symbols are used for the slope on each piece.
\rev{Each algorithm produces a symbolic representation of length $n$ using an alphabet of cardinality $k=9$. SAX and 1d-SAX requires the value of $w$ and $k$ for the reconstruction, whereas ABBA requires the $2k$ numbers representing the lengths and increments of each cluster. In total, ABBA requires more storage to represent a time series using a string of length $n$ and alphabet of cardinality $k$, but is able to represent the whole time series more accurately without truncation.}


To visualize the results of our comparison we use performance profiles \cite{DM02}. Performance profiles allow to compare the relative performance of multiple algorithms over a large set of test problems. Each algorithm is represented by a non-decreasing curve in a $\theta$--$p$ graph. The $\theta$-axis represents a tolerance $\theta\geq 1$ and the $p$-axis corresponds to a fraction $p\in [0,1]$. If a curve passes through a point $(\theta,p)$ it means that the corresponding algorithm performed within a factor~$\theta$ of the best observed performance on $100\cdot p$\,\% of the test problems. For $\theta=1$ one can read off on what fraction of all test problems each algorithm was the best performer, while as $\theta\to\infty$ all curves approach the value $p\to 1$ (unless an algorithm has failed on a fraction of the test problems, which is not the case here). 

\begin{figure}[htp!] 
\centering
\subfloat[Euclidean distance]{%
\hspace*{-5mm}\includegraphics[scale=1]{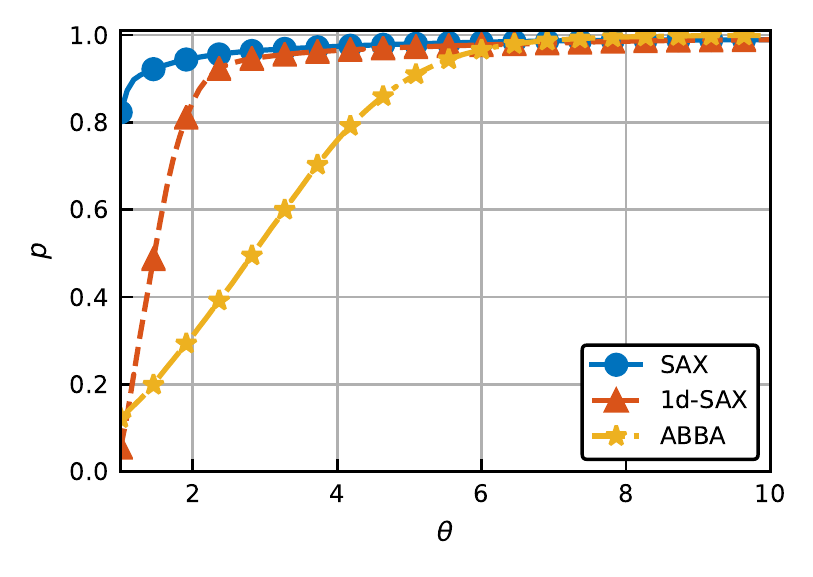}%
\label{fig:p1}%
}\hfil
\subfloat[DTW distance]{%
\hspace*{-3mm}\includegraphics[scale=1]{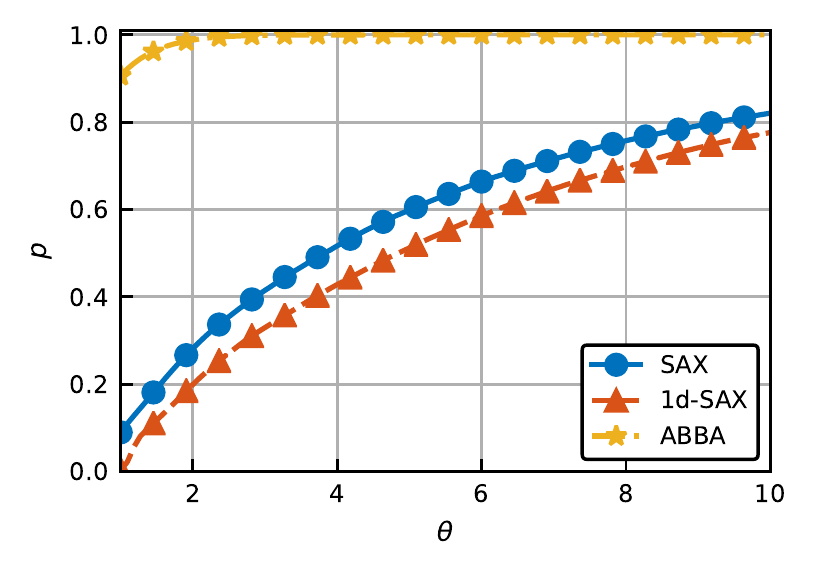}%
\label{fig:p2}%
}

\subfloat[Euclidean distance (differenced)]{%
\hspace*{-5mm}\includegraphics[scale=1]{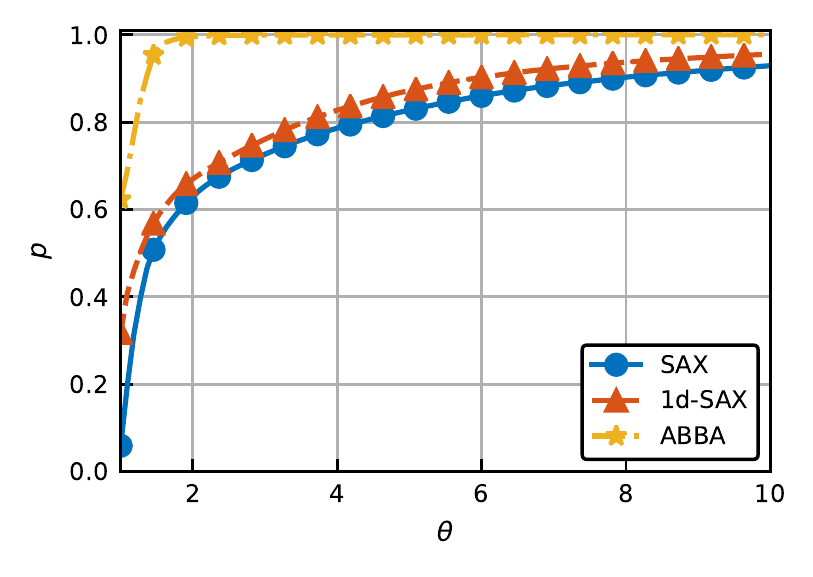}%
\label{fig:pd3}%
}\hfil
\subfloat[DTW distance (differenced)]{%
\hspace*{-3mm}\includegraphics[scale=1]{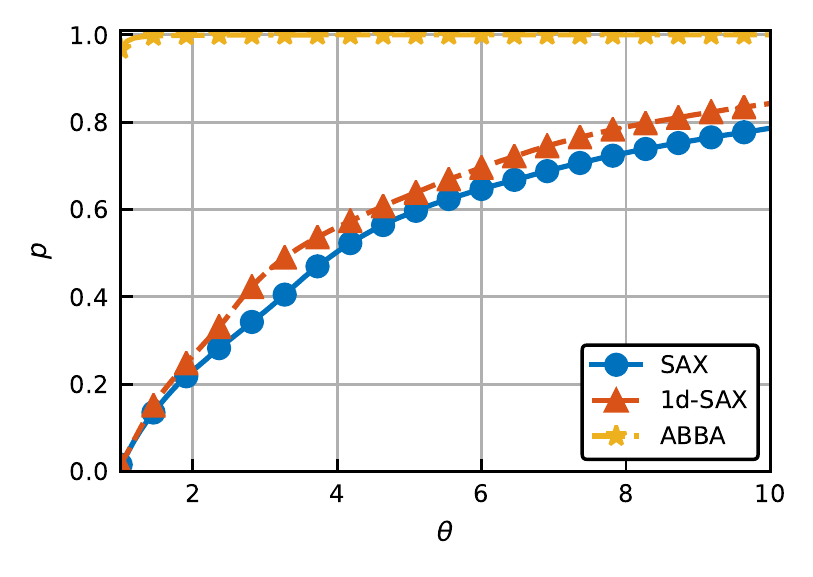}%
\label{fig:pd4}%
}
\caption{\rev{Performance profiles for the reconstruction errors of SAX, 1d-SAX, and ABBA \emph{with scaling parameter} $\texttt{scl}=0$.  Figures~\ref{fig:p1} and \ref{fig:p2} compare ABBA ($\texttt{scl}=0$) with SAX and 1d-SAX using Euclidean and Dynamic Time Warping distance, respectively.  Figures~\ref{fig:pd3} and \ref{fig:pd4} compare ABBA ($\texttt{scl}=0$) with SAX and 1d-SAX using Euclidean and Dynamic Time Warping distance of the differenced time series,  respectively.}}
\end{figure}

\begin{figure}[htp!] 
\centering
\subfloat[Euclidean distance]{%
\hspace*{-5mm}\includegraphics[scale=1]{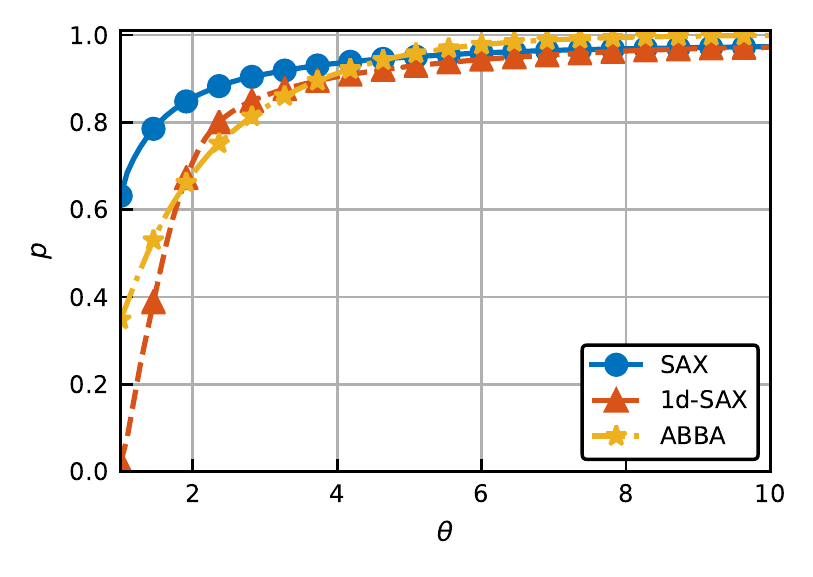}%
\label{fig:p5}%
}\hfil
\subfloat[DTW distance]{%
\hspace*{-3mm}\includegraphics[scale=1]{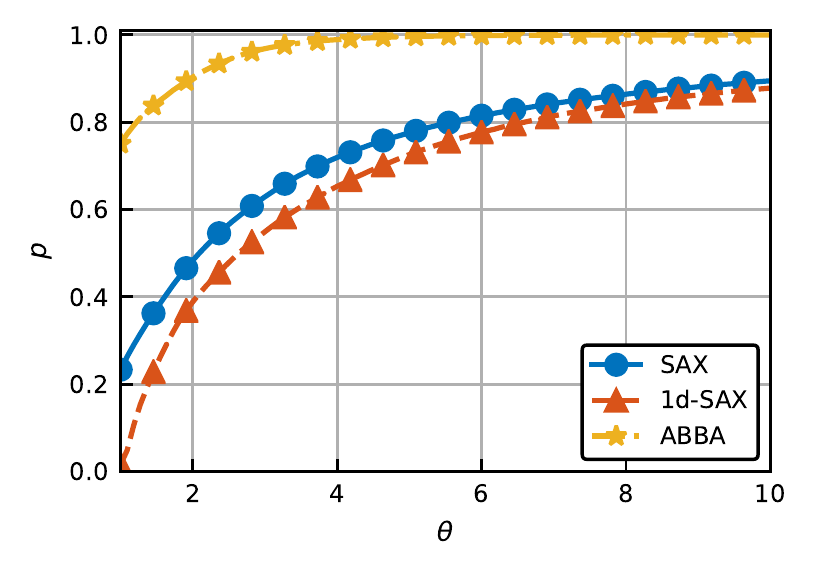}%
\label{fig:p6}%
}

\subfloat[Euclidean distance (differenced)]{%
\hspace*{-5mm}\includegraphics[scale=1]{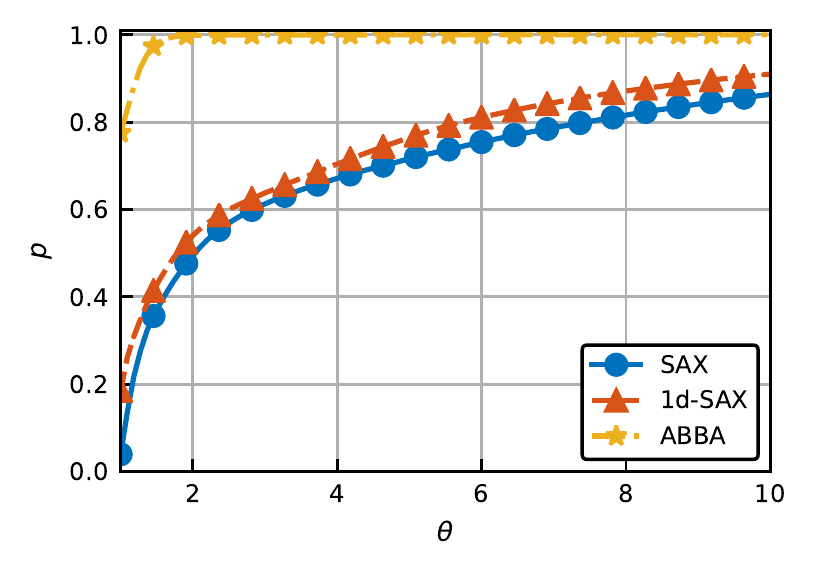}%
\label{fig:pd7}%
}\hfil
\subfloat[DTW distance (differenced)]{%
\hspace*{-3mm}\includegraphics[scale=1]{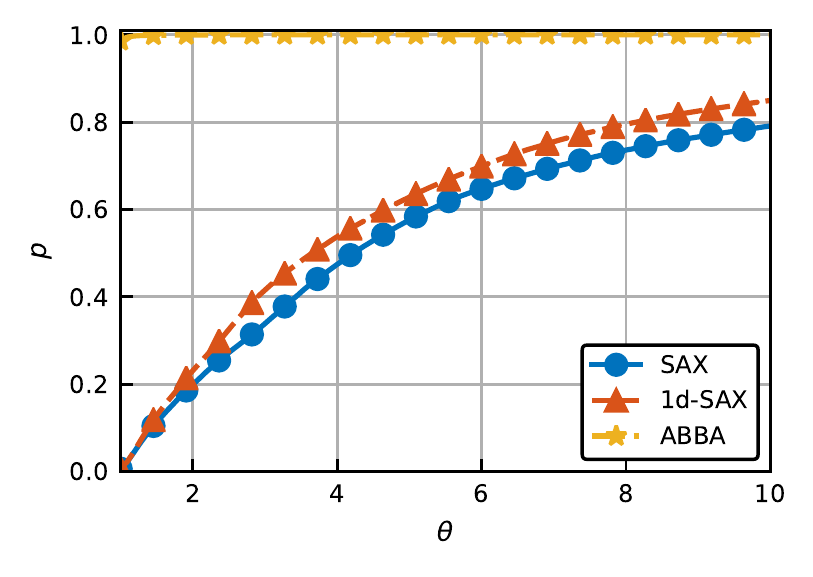}%
\label{fig:pd8}%
}

\caption{\rev{Performance profiles for the reconstruction errors of SAX, 1d-SAX, and ABBA \emph{with scaling parameter} $\texttt{scl}=1$.  Figures~\ref{fig:p5} and \ref{fig:p6} compare ABBA ($\texttt{scl}=1$) with SAX and 1d-SAX using the Euclidean and Dynamic Time Warping distance, respectively. Figures~\ref{fig:pd7} and \ref{fig:pd8} compare ABBA ($\texttt{scl}=1$) with SAX and  1d-SAX using the Euclidean and Dynamic Time Warping distance of the differenced time series,  respectively.}}
\end{figure}

In Figures~\ref{fig:p1}--\ref{fig:pd8} we present eight performance profiles for the ABBA scaling parameters $\texttt{scl} = 0$ and $\texttt{scl} = 1$, respectively, and with four different distance measures: Euclidean and DTW distances  and their differenced counterparts, respectively.
Figure~\ref{fig:p1} shows the performance profile for $\texttt{scl} = 0$, with the distance between $T$ and $\widehat T$ measured in the Euclidean norm. As expected, SAX consistently outperforms ABBA because the Euclidean distance is very sensitive to horizontal shifts in the time direction, which ABBA has completely ignored due to the $\texttt{scl} = 0$ parameter. However, it is somewhat surprising that SAX also outperforms 1d-SAX. It appears that the use of the slope information in 1d-SAX is detrimental to the approximation accuracy and, if the number of symbols is kept constant, they should better be used to represent time series values alone. This observation can also be made in the other performance profiles: irrespective of the distance measure being used, SAX with $k=9$ symbols performs better than 1d-SAX with $k=9$ symbols. 

The performance changes when we use the DTW distance, thereby allowing for shifts in time. In this case, ABBA outperforms SAX and 1d-SAX significantly; see Figure~\ref{fig:p2}. This is because ABBA has been tailored to preserve the up-and-down shape of the time series, at the cost of allowing for small errors in the lengths of the pieces which are easily  corrected by time warping. The performance gain of ABBA becomes even more pronounced when we difference the data before computing the Euclidean and DTW distances; see Figures~\ref{fig:pd3} and~\ref{fig:pd4}, respectively.

In the next four tests we set $\texttt{scl} = 1$, so the ABBA clustering procedure considers both the increments and lengths equally. Figures~\ref{fig:p5} and~\ref{fig:p6} show the resulting performance profiles using the Euclidean and DTW distance measures, respectively. As expected, ABBA becomes more competitive even for the Euclidean distance measure. Computationally, however, this comes at the cost of not being able to use a fast optimal 1d-clustering algorithm. Finally, Figures~\ref{fig:pd7} and~\ref{fig:pd8} show the performance profiles for the Euclidean and DTW distance  measures on the differenced data, respectively. As in the case $\texttt{scl} = 0$, differencing helps to improve the performance of ABBA in comparison to SAX and 1d-SAX  even further\footnote{\rev{Visual comparisons of the three algorithms on the first time series in each dataset of the UCR Time Series Classification Archive can be found at~\url{https://github.com/nla-group/ABBA/tree/master/paper/performance_profiles/scl0}.}}.

\section{\rev{Further discussion and applications} \label{sec:appl}}
\rev{Section~\ref{sec:experiment} demonstrated that ABBA provides high compression rates while guaranteeing that the time series reconstruction is still close to the original. The high compression is a consequence of the stitching procedure during the compression stage. Section~\ref{brownian_bridge} showed how errors are accumulated  piece by piece in the stitching process. We believe that this property prevents ABBA from admitting lower bounding distance measures as are available for SAX. SAX's lower bounding measure and indexability make it suitable for applications where multiple time series have to be compared (like time series classification). ABBA, on the other hand, appears best suited for applications 
where information has to be extracted from a single time series, such as anomaly detection, motif discovery, and trend prediction. As the output of ABBA is simply a string sequence, it can be combined with existing algorithms that previously used, e.g., a SAX representation. Below we discuss various aspects and applications of ABBA. 
}

\begin{figure}[htp!] 
\centering
\includegraphics[scale=1]{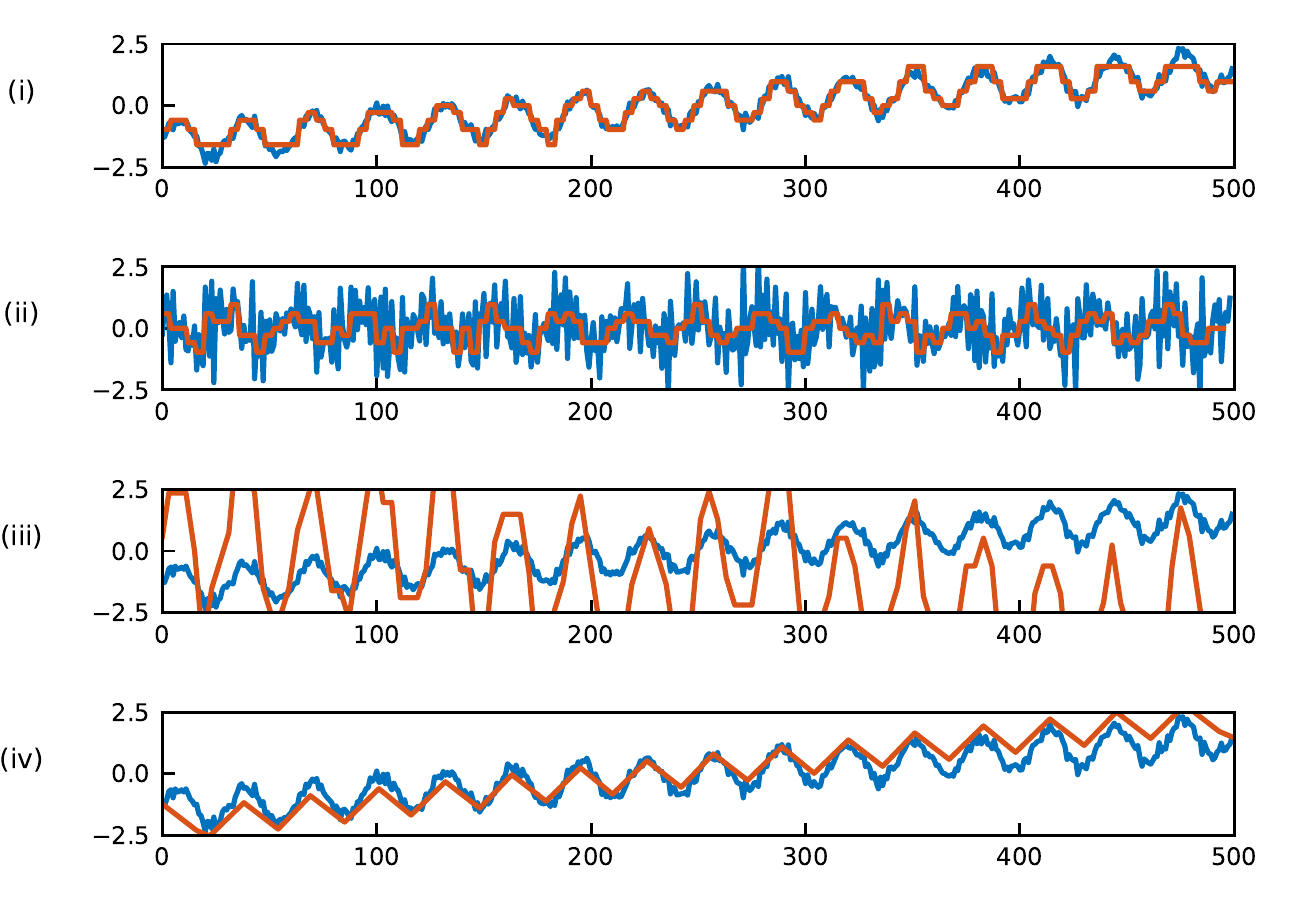}
\caption{\rev{Comparison of SAX and ABBA on a noisy sine wave with a gradual linear trend. (i) The original time series is shown in blue and the SAX representation is shown in orange. (ii) The differenced version of the original time series is shown in blue and the its SAX representation is shown in orange. (iii) The original time series is given in blue and the cumulative sum of the SAX representation from (ii) is shown in orange. (iv) The original time series is shown in blue and its ABBA representation is shown in orange.\label{fig:trend}}}
\end{figure}

\rev{\textbf{In-built differencing.} Working with the increments (instead of slopes) allows ABBA to capture linear trends in time series without preprocessing. In Figure~\ref{fig:trend} we consider the simple test problem of a sine wave with a gradual linear trend in the presence of noise. After normalization, SAX is able to accurately represent the time series as shown in Figure~\ref{fig:trend}(i). If we used the symbolic representation for trend prediction, however, the SAX representation would be unsuitable for continuing the linear trend as new symbols would need to be introduced. Of course, this problem could be overcome by removing the linear trend through  differencing the time series.
A SAX representation of the differenced time series is shown in Figure~\ref{fig:trend}(ii). Unfortunately, differencing the noisy time series amplifies the noise. Figure~\ref{fig:trend}(iii) compares the original time series against the reconstructed time series from the SAX representation of the differenced data. As we can see, the increased noise level renders the SAX representation extremely inaccurate. ABBA, on the other hand, does not require any differencing as it works with increments by default. As a consequence, the ABBA reconstruction shown in Figure~\ref{fig:trend}(iv) stays very close to the original time series, capturing both the gradual linear trend as well as the characteristic up-and-down behavior.}

\rev{\textbf{Anomaly detection} refers to the problem of finding points or intervals in time series which display surprising or unexpected behavior. Recent literature reviews of existing anomaly detection algorithms are given in \cite{GGC13, AKK18}. The ABBA representation can be used for anomaly detection in a variety of ways. Trend anomalies can be detected in the digitization procedure via $k$-means clustering of the lengths and increments. The alphabet is ordered such that \texttt{'a'} is the most frequent symbol followed by \texttt{'b'} and so forth. If the $k$th cluster contains very few elements relative to the other clusters, then this might be considered a trend anomaly.}

\rev{\textbf{TARZAN} \cite{KLC02} is a popular anomaly detection algorithm with linear time and space complexity~\cite{PFT15}. The algorithm requires two time series, a reference time series $R$ containing normal behavior and the test time series $X$. Both time series are converted to a symbolic representation and stored in a suffix tree~\cite{M76}. An anomaly score is computed by comparing the frequency of a substring in $X$ to an expected frequency computed from $R$. SAX can be used for the discretization process in TARZAN and has been shown to outperform other symbolic representations with no dimensional reduction~\cite{LKWL07}. 

If both symbolic representations are short and $X$ contains a symbol that does not appear in $R$, then the TARZAN score can suffer through lack of perspective. For example, suppose the expected frequency of the substring \texttt{'abc'} is $4.2$ and \texttt{'abc'} appears $3$~times in $X$, then the anomaly score is $3 - 4.2 = -1.2$. Suppose the symbol \texttt{'d'} does not appear in~$R$ but \texttt{'ada'} appears in $X$. The expected frequency of the substring \texttt{'ada'} is~$0$ and \texttt{'ada'} appears only once, so the anomaly score is $0 - 1 = -1$. This implies that \texttt{'abc'}  is more of an anomaly than \texttt{'ada'}. This issue can be overcome by dividing the anomaly score by the largest of the expected/actual frequency.}

\begin{figure}[htp!] 
\centering
\includegraphics[scale=1]{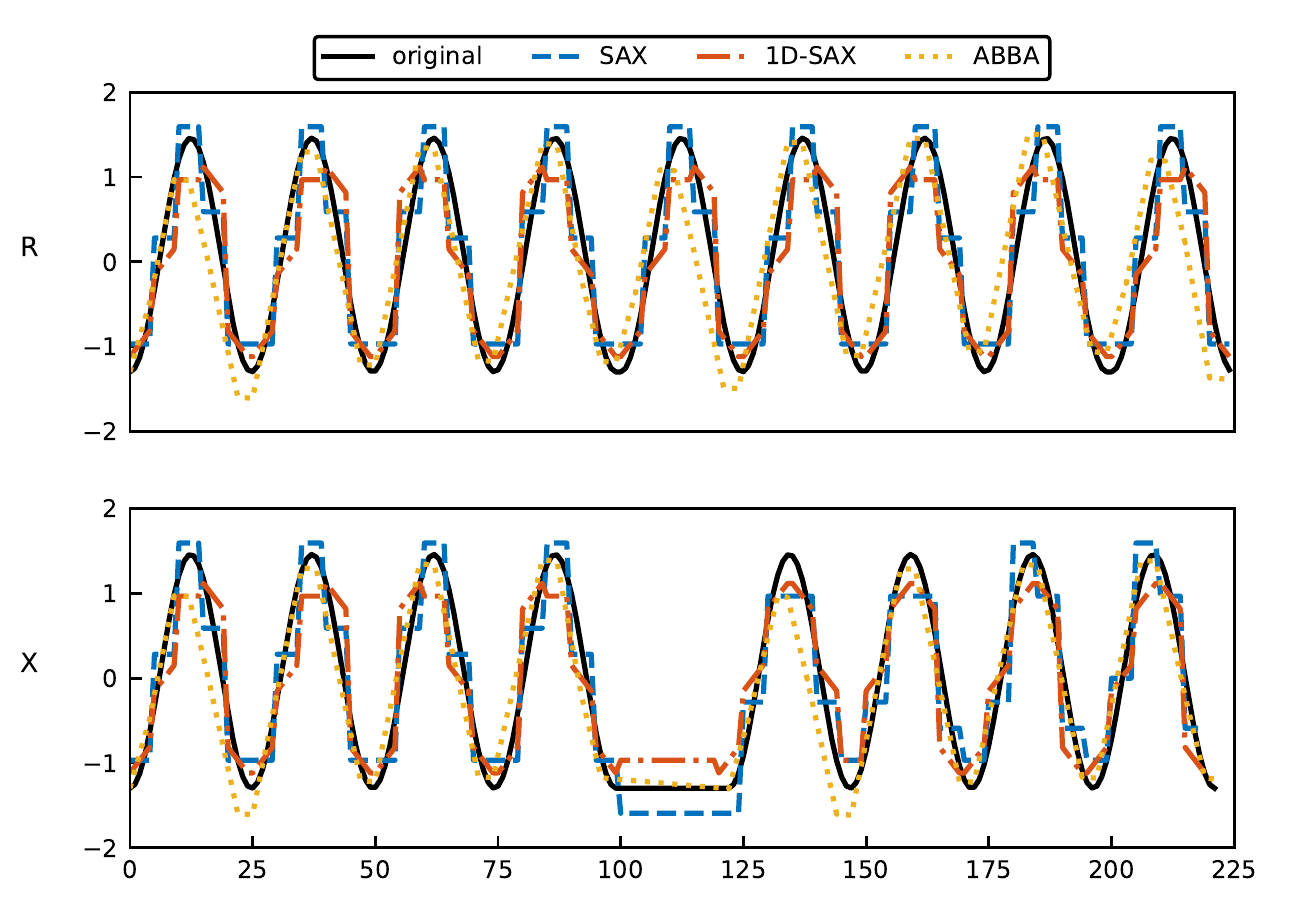}
\caption{\rev{A visual comparison of the symbolic representations of two time series. Here, $R$ is the reference time series, a simple sine wave, while $X$ is the test time series, a sine wave with a flat region slightly shorter than one wave period. \label{fig:anom_symb}}}
\end{figure}

\begin{figure}[htp!] 
\centering
\includegraphics[scale=1]{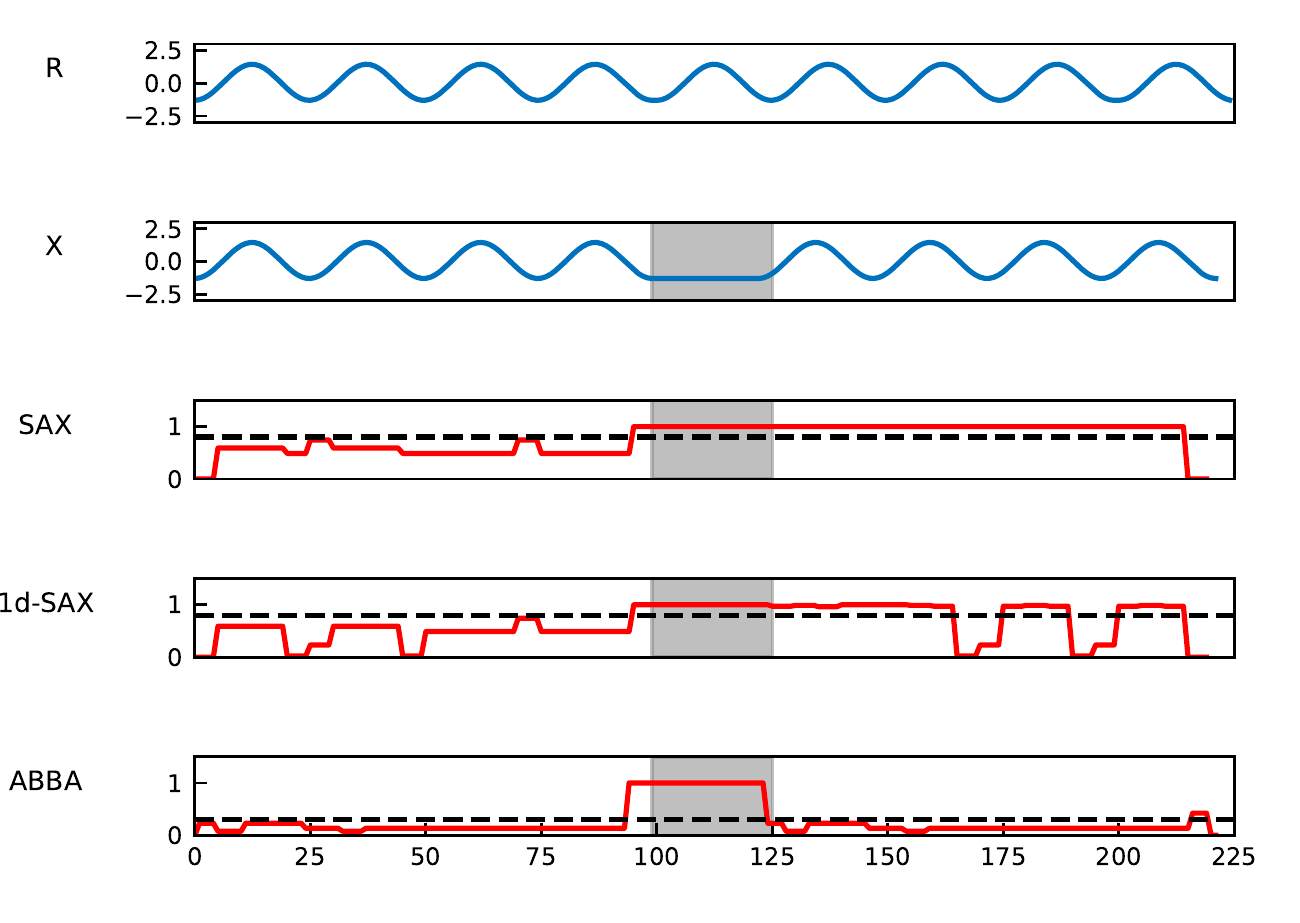}
\caption{\rev{A comparison of the TARZAN anomaly detection algorithm using the SAX, 1d-SAX, and ABBA representations, respectively. The first time series $R$ is the reference, while the second time series $X$ is to be tested. The final three plots show the adapted TARZAN anomaly scores for the SAX, 1d-SAX, and ABBA representations,  respectively. The black dashed lines indicate   tolerances that could be used define the anomalies.\label{fig:anom_tarzan}}}
\end{figure}

\rev{In Figures~\ref{fig:anom_symb} and \ref{fig:anom_tarzan} we consider a simple experiment comparing SAX, 1d-SAX, and ABBA as discretization procedures for TARZAN with the modified anomaly score\footnote{\rev{A Python implementation of TARZAN which supports the use of SAX, 1d-SAX, and ABBA can be downloaded from\\  \url{https://github.com/nla-group/TARZAN}.}}. The reference time series $R$ is a simple sine wave where each period spans $25$ time samples. The time series $X$ has a full wave replaced by a flat line of $22$ time points. The SAX and 1d-SAX representations use a window length $w=5$ and $k=9$ symbols,  whereas ABBA uses a tolerance tuned to give a symbolic representation of equal length and $k$ is bounded by $9$. The time series $R$ and $X$ and their symbolic reconstructions are shown in Figure~\ref{fig:anom_symb}. If the length of the anomaly does not align with the window length $w$, then SAX and 1d-SAX tend to represent the sine wave following the anomaly as a different substring. The adapted TARZAN score is required as certain symbols appear in $X$ that do not appear in~$R$. Figure~\ref{fig:anom_tarzan} shows the resulting TARZAN anomaly scores. Both SAX and 1d-SAX suffer from the fixed window length, returning high anomaly scores throughout time following the anomaly, whereas TARZAN using ABBA is able to recover almost immediately after the anomaly due to the adaptive segment lengths.}
 
\rev{\textbf{VizTree.} We finally mention the possibility of representing an ABBA output as a VizTree, a time series pattern discovery and visualization tool based on suffix trees \cite{LKL04a, LKL04b, LKL05}. The authors use SAX to discretize the time series before building a suffix tree. Each branch of the suffix tree represents a substring and the thickness of that branch represents the frequency of the substring in the symbolic representation. In principle, SAX pairs well with the visualization as the Gaussian breakpoints should ensure that each symbol appears equally likely. In practice, this is often not the case. One could use ABBA's discretization process instead of SAX by relating the thickness of each line to the frequency of the symbols determined in the clustering procedure. A poor choice of the window length~$w$ in the piecewise aggregate approximation in SAX could lead to missing motifs if the distance between is not near a multiple of~$w$. Furthermore, SAX might fail to detect motifs if time warping has occurred, whilst VizTree via ABBA should be able to better capture time-warped motifs as the segment lengths are chosen adaptively. A further exploration of this application will be the subject of future work.}

\section{Conclusions and future work}\label{sec:conc}
We introduced ABBA, an adaptive symbolic time series representation which aims to preserve the essential shape of a time series. We have shown that the ABBA representation has favorable approximation properties compared to other popular representations, in particular, when the dynamic time warping distance is used. \rev{Furthermore, we demonstrated the use of ABBA in some important data mining applications, including trend prediction and anomaly detection.} Future research will be devoted to an online streaming version of ABBA with the necessary adaptations of the the Brownian bridge-based error analysis, \rev{as well as a more in-depth study of VizTree visualizations. Our recent work \cite{EG20} explores ABBA's potential for time series forecasting.} 

\begin{acknowledgements}
This work was supported by the Engineering and Physical Sciences Research Council (EPRSC), grant EP/N509565/1. We thank Sabisu and EPSRC for providing SE with a CASE PhD studentship. SG acknowledges support from the Alan Turing Institute. We thank Timothy D.~Butters for his help with C++ and SWIG, and are grateful to  Eamonn Keogh and all other contributors to the UCR Time Series Classification Archive. \rev{We also thank the three anonymous referees and the editor for their helpful comments which significantly improved the paper.}
\end{acknowledgements}

\bibliographystyle{mysiam}      

\bibliography{references_paper}

\begin{thebibliography}{10}

\bibitem{AML18}
{\sc A.~Abanda, U.~Mori, and J.~A. Lozano},
  \href{https://doi.org/10.1007/s10618-018-0596-4}{{\em A review on distance
  based time series classification}}, Data Min. Knowl. Discov., 33 (2019),
  pp.~378--412.

\bibitem{ASW15}
{\sc S.~Aghabozorgi, A.~S. Shirkhorshidi, and T.~Y. Wah},
  \href{http://www.sciencedirect.com/science/article/pii/S0306437915000733}{{\em
  Time-series clustering--a decade review}}, Inf. Syst., 53 (2015), pp.~16--38.

\bibitem{AV06}
{\sc D.~Arthur and S.~Vassilvitskii},
  \href{http://theory.stanford.edu/~sergei/papers/kMeans-socg.pdf}{{\em How
  slow is the {$k$}-means method?}}, in Symposium on Computational Geometry,
  vol.~6, ACM, New York, 2006, pp.~1--10.

\bibitem{AKK18}
{\sc G.~Atluri, A.~Karpatne, and V.~Kumar},
  \href{https://doi.org/10.1145/3161602}{{\em Spatio-temporal data mining: A
  survey of problems and methods}}, ACM Comput. Surv., 51 (2018), p.~83.

\bibitem{BBO12}
{\sc P.~M. Barnaghi, A.~A. Bakar, and Z.~A. Othman},
  \href{http://dx.doi.org/10.3923/ijscomp.2013.261.268}{{\em Enhanced symbolic
  aggregate approximation method for financial time series data
  representation}}, in Proceedings of the 6th International Conference on New
  Trends in Information Science, Service Science and Data Mining, IEEE, 2012,
  pp.~790--795.

\bibitem{BR15}
{\sc M.~G. Baydogan and G.~Runger},
  \href{https://doi.org/10.1007/s10618-014-0349-y}{{\em Learning a symbolic
  representation for multivariate time series classification}}, Data Min.
  Knowl. Discov., 29 (2015), pp.~400--422.

\bibitem{BBH15}
{\sc Y.~Benyahmed, A.~A. Bakar, A.~R. Hamdan, and S.~M.~S. Abdullah},
  \href{https://pdfs.semanticscholar.org/4b7e/2914e0782fa28e3fd01a66bd32c1b563829b.pdf}{{\em
  A time-weighted average-based {PAA} representation for time series
  symbolization}}, International Journal of Advances in Soft Computing \& Its
  Applications,  (2015).

\bibitem{BC94}
{\sc D.~J. Berndt and J.~Clifford},
  \href{https://www.aaai.org/Papers/Workshops/1994/WS-94-03/WS94-03-031.pdf}{{\em
  Using dynamic time warping to find patterns in time series}}, in KDD
  Workshop, vol.~10, 1994, pp.~359--370.

\bibitem{BR14}
{\sc V.~Bettaiah and H.~S. Ranganath},
  \href{http://doi.acm.org/10.1145/2638404.2638475}{{\em An analysis of time
  series representation methods: data mining applications perspective}}, in
  Proceedings of the ACM Southeast Regional Conference, ACM, 2014,
  pp.~16:1--16:6.

\bibitem{BBC16}
{\sc A.~Bondu, M.~Boull{\'e}, and A.~Cornu{\'e}jols},
  \href{https://www.researchgate.net/publication/283535717_Symbolic_representation_of_time_series_A_hierarchical_coclustering_formalization}{{\em
  Symbolic representation of time series: a hierarchical coclustering
  formalization}}, in International Workshop on Advanced Analysis and Learning
  on Temporal Data, Springer, 2016, pp.~3--16.

\bibitem{B01}
{\sc M.~Boull{\'e}}, \href{https://doi.org/10.1007/s10994-006-8364-x}{{\em
  {MODL}: A bayes optimal discretization method for continuous attributes}},
  Mach. Learn., 65 (2006), pp.~131--165.

\bibitem{CKL03}
{\sc B.~Chiu, E.~Keogh, and S.~Lonardi},
  \href{https://doi.org/10.1145/956750.956808}{{\em Probabilistic discovery of
  time series motifs}}, in Proceedings of the 9th International Conference on
  Knowledge Discovery and Data Mining, ACM, 2003, pp.~493--498.

\bibitem{UCRArchive}
{\sc H.~A. Dau, E.~Keogh, K.~Kamgar, C.-C.~M. Yeh, Y.~Zhu, S.~Gharghabi, C.~A.
  Ratanamahatana, Yanping, B.~Hu, N.~Begum, A.~Bagnall, A.~Mueen, and
  G.~Batista},
  \href{https://www.cs.ucr.edu/~eamonn/time_series_data_2018/}{{\em The {UCR}
  time series classification archive}}, October 2018.

\bibitem{DM02}
{\sc E.~D. Dolan and J.~J. Mor\'{e}},
  \href{https://doi.org/10.1007/s101070100263}{{\em Benchmarking optimization
  software with performance profiles}}, Math. Program., 91 (2002),
  pp.~201--213.

\bibitem{EG20}
{\sc S.~Elsworth and S.~G\"{u}ttel},
  \href{https://arxiv.org/abs/2003.05672}{{\em {Time Series Forecasting Using
  LSTM Networks: A Symbolic Approach}}}, arXiv EPrint arXiv:2003.05672v1,
  Manchester Institute for Mathematical Sciences, The University of Manchester,
  UK, 2020.

\bibitem{EART12}
{\sc B.~Esmael, A.~Arnaout, R.~K. Fruhwirth, and G.~Thonhauser},
  \href{https://doi.org/10.1007/978-3-642-31128-4_29}{{\em Multivariate time
  series classification by combining trend-based and value-based
  approximations}}, in Proceedings of the International Conference on
  Computational Science and Its Applications, Springer, 2012, pp.~392--403.

\bibitem{FU11}
{\sc T.-c. Fu}, \href{http://dx.doi.org/10.1016/j.engappai.2010.09.007}{{\em A
  review on time series data mining}}, Eng. Appl. Artif. Intell., 24 (2011),
  pp.~164--181.

\bibitem{GBC13}
{\sc F.~{Ganz}, P.~{Barnaghi}, and F.~{Carrez}},
  \href{http://dx.doi.org/10.1109/JSEN.2013.2271562}{{\em Information
  abstraction for heterogeneous real world internet data}}, IEEE Sensors J., 13
  (2013), pp.~3793--3805.

\bibitem{GWS14}
{\sc J.~Grabocka, M.~Wistuba, and L.~Schmidt-Thieme},
  \href{http://dx.doi.org/10.1109/TKDE.2014.2377746}{{\em Scalable
  classification of repetitive time series through frequencies of local
  polynomials}}, IEEE Trans. Knowl. Data Eng., 27 (2014), pp.~1683--1695.

\bibitem{GLMN17}
{\sc A.~Gr{\o}nlund, K.~G. Larsen, A.~Mathiasen, and J.~S. Nielsen},
  \href{http://arxiv.org/abs/1701.07204}{{\em Fast exact k-means, k-medians and
  {B}regman divergence clustering in 1{D}}}, arXiv preprint arXiv:1701.07204,
  (2017).

\bibitem{GGC13}
{\sc M.~Gupta, J.~Gao, C.~C. Aggarwal, and J.~Han},
  \href{http://dx.doi.org/10.1109/TKDE.2013.184}{{\em Outlier detection for
  temporal data: A survey}}, IEEE Trans. Knowl. Data Eng., 26 (2013),
  pp.~2250--2267.

\bibitem{KCHP01}
{\sc E.~Keogh, S.~Chu, D.~Hart, and M.~Pazzani},
  \href{http://dx.doi.org/10.1109/ICDM.2001.989531}{{\em An online algorithm
  for segmenting time series}}, in Proceedings of the 2001 {IEEE} International
  Conference on Data Mining, 2001, pp.~289--296.

\bibitem{KK03}
{\sc E.~Keogh and S.~Kasetty},
  \href{https://doi.org/10.1023/A:1024988512476}{{\em On the need for time
  series data mining benchmarks: a survey and empirical demonstration}}, Data
  Min. Knowl. Discov., 7 (2003), pp.~349--371.

\bibitem{KLC02}
{\sc E.~Keogh, S.~Lonardi, and B.-c. Chiu},
  \href{https://doi.org/10.1145/775047.775128}{{\em Finding surprising patterns
  in a time series database in linear time and space}}, in Proceedings of the
  8th International Conference on Knowledge Discovery and Data Mining, ACM,
  2002, pp.~550--556.

\bibitem{KR05}
{\sc E.~Keogh and C.~A. Ratanamahatana},
  \href{https://doi.org/10.1007/s10115-004-0154-9}{{\em Exact indexing of
  dynamic time warping}}, Knowl. Inf. Syst., 7 (2005), pp.~358--386.

\bibitem{KP01}
{\sc E.~J. Keogh and M.~J. Pazzani},
  \href{https://www.ics.uci.edu/~pazzani/Publications/sdm01.pdf}{{\em
  Derivative dynamic time warping}}, in Proceedings of the 2001 {SIAM}
  International Conference on Data Mining, 2001, pp.~1--11.

\bibitem{K00}
{\sc J.-Y. Kim},
  \href{http://www.sciencedirect.com/science/article/pii/S0304407699000317}{{\em
  Detection of change in persistence of a linear time series}}, J.
  Econometrics, 95 (2000), pp.~97--116.

\bibitem{LZY12}
{\sc G.~Li, L.~Zhang, and L.~Yang},
  \href{http://dx.doi.org/10.1007/978-3-642-32695-0_25}{{\em {TSX}: a novel
  symbolic representation for financial time series}}, in PRICAI 2012: Trends
  in Artificial Intelligence, Springer, 2012, pp.~262--273.

\bibitem{LKL05}
{\sc J.~Lin, E.~Keogh, and S.~Lonardi},
  \href{https://doi.org/10.1057/palgrave.ivs.9500089}{{\em Visualizing and
  discovering non-trivial patterns in large time series databases}}, Inform.
  Visual., 4 (2005), pp.~61--82.

\bibitem{LKL04a}
{\sc J.~Lin, E.~Keogh, S.~Lonardi, J.~P. Lankford, and D.~M. Nystrom},
  \href{https://doi.org/10.1145/1014052.1014104}{{\em Visually mining and
  monitoring massive time series}}, in Proceedings of the 10th International
  Conference on Knowledge Discovery and Data Mining, ACM, 2004, pp.~460--469.

\bibitem{LKL04b}
{\sc J.~Lin, E.~Keogh, S.~Lonardi, J.~P. Lankford, and D.~M. Nystrom},
  \href{http://dx.doi.org/10.1016/B978-012088469-8.50124-8}{{\em Viztree: a
  tool for visually mining and monitoring massive time series databases}}, in
  Proceedings of the 30th International Conference on Very large Data Bases,
  VLDB Endowment, 2004, pp.~1269--1272.

\bibitem{LKWL07}
{\sc J.~Lin, E.~Keogh, L.~Wei, and S.~Lonardi},
  \href{https://doi.org/10.1007/s10618-007-0064-z}{{\em Experiencing {SAX}: a
  novel symbolic representation of time series}}, Data Min. Knowl. Discov., 15
  (2007), pp.~107--144.

\bibitem{LSK06}
{\sc B.~Lkhagva, Y.~Suzuki, and K.~Kawagoe}, \href{http://dx.doi.org/doi =
  {10.1109/ICDEW.2006.99}}{{\em New time series data representation {ESAX} for
  financial applications}}, in Proceedings of the 22nd International Conference
  on Data Engineering Workshops, 2006, pp.~x115--x115.

\bibitem{LYCLFHM15}
{\sc G.~Luo, K.~Yi, S.-W. Cheng, Z.~Li, W.~Fan, C.~He, and Y.~Mu},
  \href{http://dx.doi.org/10.1109/ICDE.2015.7113282}{{\em Piecewise linear
  approximation of streaming time series data with max-error guarantees}}, in
  Proceedings of the {IEEE} 31st International Conference on Data Engineering,
  2015, pp.~173--184.

\bibitem{MGQT13}
{\sc S.~Malinowski, T.~Guyet, R.~Quiniou, and R.~Tavenard},
  \href{http://dx.doi.org/10.1007/978-3-642-41398-8_24}{{\em 1d-{SAX}: a novel
  symbolic representation for time series}}, in Advances in Intelligent Data
  Analysis XII, Springer, 2013, pp.~273--284.

\bibitem{M76}
{\sc E.~M. McCreight}, \href{https://doi.org/10.1145/321941.321946}{{\em A
  space-economical suffix tree construction algorithm}}, J. ACM, 23 (1976),
  pp.~262--272.

\bibitem{MU06}
{\sc F.~M{\"o}rchen and A.~Ultsch},
  \href{http://dx.doi.org/10.1007/3-540-31314-1_33}{{\em Finding persisting
  states for knowledge discovery in time series}}, in From Data and Information
  Analysis to Knowledge Engineering, Springer, 2006, pp.~278--285.

\bibitem{scikit}
{\sc F.~Pedregosa, G.~Varoquaux, A.~Gramfort, V.~Michel, B.~Thirion, O.~Grisel,
  M.~Blondel, P.~Prettenhofer, R.~Weiss, V.~Dubourg, J.~Vanderplas, A.~Passos,
  D.~Cournapeau, M.~Brucher, M.~Perrot, and E.~Duchesnay},
  \href{http://dx.doi.org/10.5555/1953048.2078195}{{\em Scikit-learn: machine
  learning in {P}ython}}, J. Mach. Learn. Res., 12 (2011), pp.~2825--2830.

\bibitem{PFT15}
{\sc T.~Pelkonen, S.~Franklin, J.~Teller, P.~Cavallaro, Q.~Huang, J.~Meza, and
  K.~Veeraraghavan}, \href{https://doi.org/10.14778/2824032.2824078}{{\em
  Gorilla: A fast, scalable, in-memory time series database}}, in Proceedings
  of the VLDB Endowment, vol.~8, VLDB Endowment, 2015, pp.~1816--1827.

\bibitem{PLD10}
{\sc N.~D. Pham, Q.~L. Le, and T.~K. Dang},
  \href{http://dl.acm.org/citation.cfm?id=1894753.1894768}{{\em {HOT} a{SAX}: a
  novel adaptive symbolic representation for time series discords discovery}},
  in Proceedings of the 2nd International Conference on Intelligent Information
  and Database Systems, Springer, 2010, pp.~113--121.

\bibitem{SK08}
{\sc J.~Shieh and E.~Keogh},
  \href{https://www.vldb.org/pvldb/vol8/p1816-teller.pdf}{{\em i{SAX}: indexing
  and mining terabyte sized time series}}, in Proceedings of the 14th
  International Conference on Knowledge Discovery and Data Mining, ACM, 2008,
  pp.~623--631.

\bibitem{WS11}
{\sc H.~Wang and M.~Song},
  \href{https://journal.r-project.org/archive/2011-2/RJournal_2011-2_Wang+Song.pdf}{{\em
  Ckmeans.1d.dp: optimal $k$-means clustering in one dimension by dynamic
  programming}}, R J., 3 (2011), pp.~29--33.

\bibitem{ZLCH18}
{\sc K.~Zhang, Y.~Li, Y.~Chai, and L.~Huang},
  \href{http://dx.doi.org/10.1109/CCDC.2018.8407498}{{\em Trend-based symbolic
  aggregate approximation for time series representation}}, in Proceedings of
  the Chinese Control And Decision Conference, IEEE, 2018, pp.~2234--2240.

\end{thebibliography}
\end{document}